\documentclass{article}

\usepackage{iclr2026_conference}
\iclrfinalcopy
\usepackage{graphicx}
\usepackage{amsmath}
\usepackage{amssymb}
\usepackage{amsthm}
\usepackage{mathtools}
\usepackage{microtype}
\usepackage{booktabs}
\usepackage{enumitem}
\usepackage{pifont}
\usepackage{algorithm}
\usepackage{algorithmic}
\usepackage{adjustbox}
\usepackage{comment}
\usepackage{titletoc}
\usepackage{tikz}
\usepackage{pgfplots}
\pgfplotsset{compat=1.18}
\usepgfplotslibrary{polar}
\usetikzlibrary{positioning}
\usepackage{subcaption}
\usepackage[colorlinks=true,linkcolor=blue,citecolor=blue,urlcolor=blue,breaklinks=true]{hyperref}

\newcommand{\etal}{\textit{et al.}}

\providecommand{\keywords}[1]{}

\title{Distilling Specialized Orders for Visual Generation}

\author{
Rishav Pramanik$^{1}$, Amin Sghaier$^{2,3,4}$, Masih Aminbeidokhti$^{2,3}$, Juan A. Rodriguez$^{2,4,5}$,\And Antoine Poupon$^{2,6}$,
David Vazquez$^{5}$, Christopher Pal$^{4,5,7,8}$, Zhaozheng Yin$^{1}$, Marco Pedersoli$^{2,3}$\\[6pt]
$^{1}$Stony Brook University, NY, USA\\
$^{2}$International Laboratory on Learning Systems (ILLS)\\
$^{3}$LIVIA, \'{E}TS Montr\'{e}al, QC, Canada\\
$^{4}$Mila--Quebec AI Institute\\
$^{5}$ServiceNow Research\\
$^{6}$Universit\'{e} Paris-Saclay, CentraleSup\'{e}lec, France\\
$^{7}$Polytechnique Montr\'{e}al\\
$^{8}$Canada CIFAR AI Chair
}

\begin{document}

\maketitle

\begin{abstract}
Autoregressive (AR) image generators are becoming increasingly popular due to their ability to produce high-quality images and their scalability. Typical AR models are locked onto a specific generation order, often a raster-scan from top-left to bottom-right; this prohibits multi-task flexibility (inpainting, editing, outpainting) without retraining. Any-order AR models address this by learning to generate under arbitrary patch orderings, but at the cost of increased complexity and lower performance. In this paper, we present Ordered Autoregressive (OAR) generation, a self-distillation pipeline that first trains an any-order AR model, then extracts specialized generation orders from the model's own confidence scores, and fine-tunes on these orders. This achieves two goals: 1) improved generation quality by redirecting capacity from learning all $N!$ orderings to a single specialized path, and 2) preserved flexibility of any-order models. On ImageNet $256\times 256$, OAR improves FID from 2.39 to 2.17 over the any-order baseline, with consistent gains on Fashion Products and CelebA-HQ. OAR supports zero-shot inpainting and outpainting without retraining, and human evaluation shows 64\% preference over the baseline. The pipeline requires only lightweight fine-tuning on a pretrained any-order model, with no architectural changes or additional annotations.


\begin{figure*}[t]
    \centering
    \hspace{-0.6cm}
    \begin{subfigure}[t]{0.33\textwidth}
        \centering
        \textbf{(a) Generation}\\[3pt]
        \begin{tabular}{@{}c@{\hskip 3pt}c@{}}
            \includegraphics[width=0.31\linewidth]{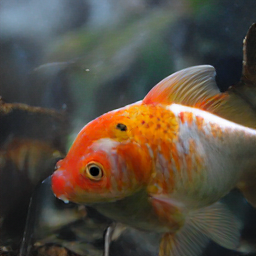} &
            \includegraphics[width=0.31\linewidth]{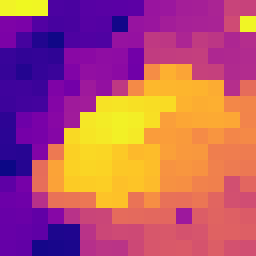} \\
            \scriptsize Image & \scriptsize Order \\[6pt]
            \includegraphics[width=0.31\linewidth]{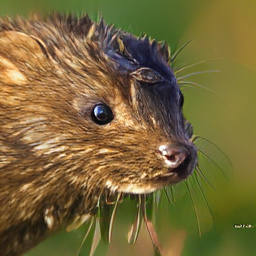} &
            \includegraphics[width=0.31\linewidth]{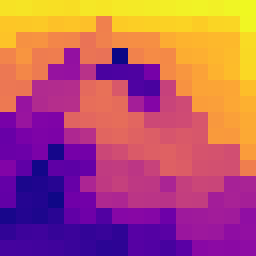} \\
            \scriptsize Image & \scriptsize Order
        \end{tabular}
    \end{subfigure}
    \hspace{-.1cm}
    \begin{subfigure}[t]{0.33\textwidth}
        \centering
        \textbf{(b) Class-Conditional Editing}\\[3pt]
        \begin{tabular}{@{}c@{\hskip 2pt}c@{\hskip 2pt}c@{}}
            \includegraphics[width=0.31\linewidth]{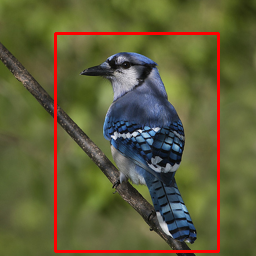} &
            \includegraphics[width=0.31\linewidth]{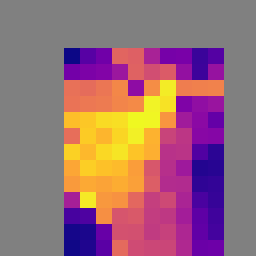} &
            \includegraphics[width=0.31\linewidth]{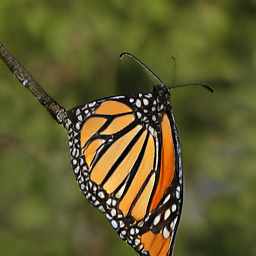} \\
            \scriptsize Masked & \scriptsize Order & \scriptsize Monarch 
            \\[6pt]
            \includegraphics[width=0.31\linewidth]{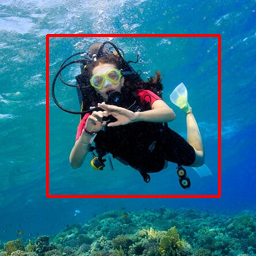} &
            \includegraphics[width=0.31\linewidth]{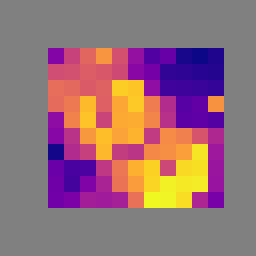} &
            \includegraphics[width=0.31\linewidth]{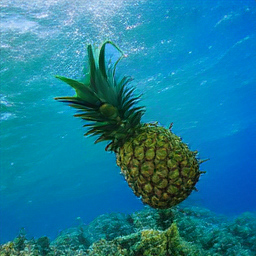} \\
            \scriptsize Masked & \scriptsize Order & \scriptsize Pineapple 
        \end{tabular}
    \end{subfigure}%
    \hfill
    \begin{subfigure}[t]{0.33\textwidth}
        \centering
        \textbf{(c) Inpainting}\\[3pt]
        \begin{tabular}{@{}c@{\hskip 2pt}c@{\hskip 2pt}c@{}}
            \includegraphics[width=0.31\linewidth]{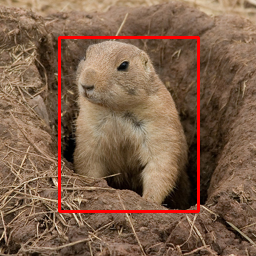} &
            \includegraphics[width=0.31\linewidth]{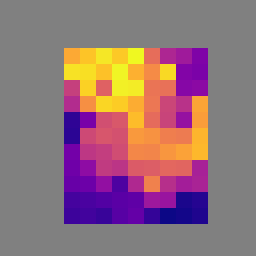} &
            \includegraphics[width=0.31\linewidth]{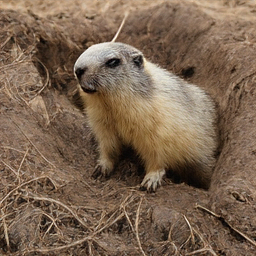} \\
            \scriptsize Masked & \scriptsize Order & \scriptsize Marmot 
            \\[6pt]
            \includegraphics[width=0.31\linewidth]{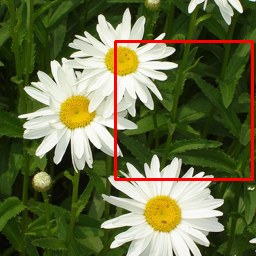} &
            \includegraphics[width=0.31\linewidth]{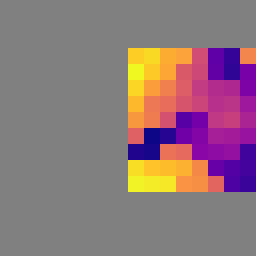} &
            \includegraphics[width=0.31\linewidth]{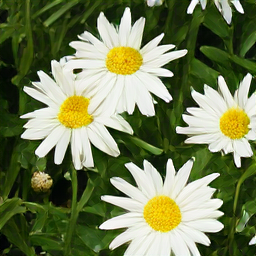} \\
            \scriptsize Masked & \scriptsize Order & \scriptsize Daisy 
        \end{tabular}
    \end{subfigure}
    \caption{\textbf{Visualizing Generation and Zero-Shot Capabilities.} We demonstrate the flexibility of our model on both generation and tasks it was not explicitly trained for. \textbf{(a) Generation:} Class-conditional image generation results with corresponding specialized orders. \textbf{(b) Editing:} Transforming objects into different classes. Note how the order (middle) adapts to the new content. \textbf{(c) Inpainting:} Same-class generation. The model can perform coherent reconstruction (Marmot) or completion (Daisy), where it generates a new flower.}
    \label{fig:inpainting_editing}
\end{figure*}
\end{abstract}

\section{Introduction}
\label{sec:intro}


Autoregressive (AR) models have become the de-facto choice for state-of-the-art generative language models~\cite{brown2020language}, underpinning modern LLMs~\cite{openai2024gpt4technicalreport, geminiteam2024, yang2024qwen2technicalreport, yang2025qwen3} that demonstrate remarkable versatility across diverse tasks. A mature ecosystem of optimizations: KV caching, pipeline and tensor parallelism, efficient attention methods has made AR models highly scalable in practice~\cite{TheC3, dao2024flashattention, kwon2023efficient}. Naturally, this success has motivated the adoption of AR models for image generation~\cite{ramesh2021zero, esser2021taming, sun2024autoregressivemodelbeatsdiffusion}. Typical AR image generators remain largely single-purpose, failing to natively support diverse tasks like inpainting, outpainting, or region-based editing without task-specific retraining. This functional stagnation is rooted in a fundamental architectural constraint: the commitment to a fixed raster-scan generation order. By treating the generation order as an immutable recipe rather than a steerable degree of freedom, current models are structurally locked into a single generation trajectory. In this paradigm, the generation order is not leveraged as a functional mechanism for task specification; instead, it is imposed as a rigid inductive bias that prohibits the very versatility and spatial literacy required for general-purpose vision.



The underlying principle which enables AR models to generate complex data is the factorization of the joint probability of the data into a chain of conditional probabilities. Learning these conditional probabilities is the core of AR modeling and also makes the learning process more tractable than directly modeling the full joint distribution~\cite{shih2022training, chen2018pixelsnail}. 
In standard AR image generation, this factorization is fixed to a single permutation of non overlapping patches in a raster-scan order. Training with multiple permutations of the patches allows to predict any patch conditioned on any subset of other patches, regarless of their spatial arrangement. With this capability, the same model can be further used on multiple tasks without retraining, such as inpainting, outpainting, editing, and completion without explicitly training for these tasks.



On the other hand, this quality comes at a cost of using up the model capacity to learn all possible chain of conditionals. In theory, the joint distribution should be able to factorized into any ordering of the data sequencel. In practice this is not the case, mostly due to the fact that it involves iteratively training over $N!$ permutations of the data sequence, which is computationally infeasible. As a result of this, conditional probabilities are estimated, different data orderings can produce substantially different outcomes~\cite{vinyals2016order}. This insight motivates our work: \textit{Can a single autoregressive model, trained to generate in any order, both unlock flexible generation capabilities (inpainting, editing) and discover specialized orders that improve generation quality?}



\begin{figure}
    \centering
    \includegraphics[width=0.7
\linewidth,keepaspectratio]{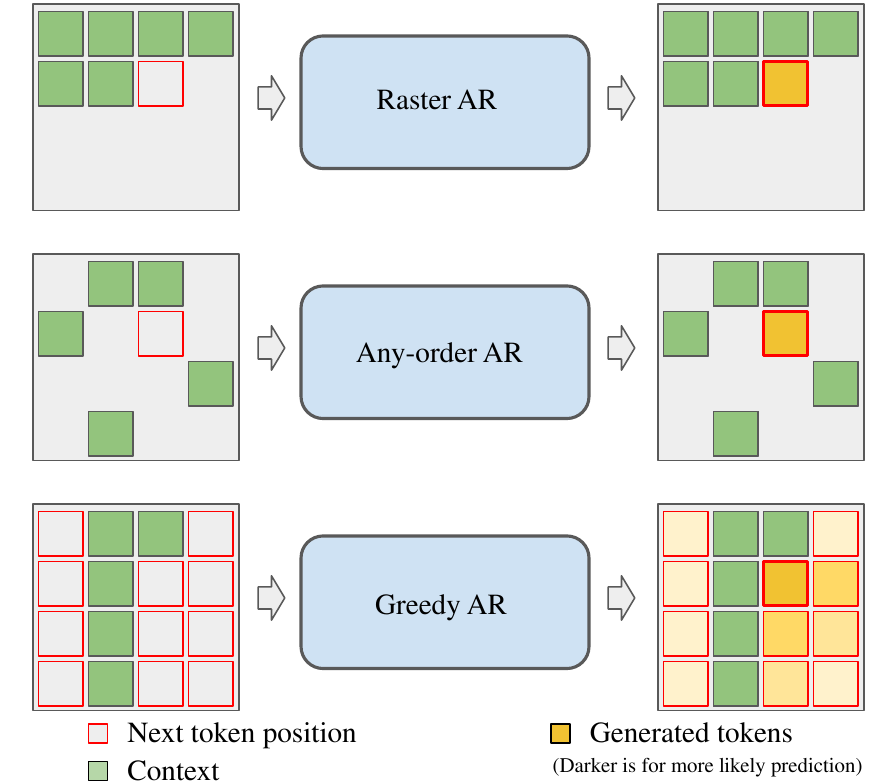}
    \caption{Different Autoregressive (AR) image generation models. (\textbf{Top}) A traditional AR is the normal approach for autoregressive generation from top left to bottom-right. The input token contains the content $x_i$ and the position $l_i$. (\textbf{Middle}) Any-order AR learns to generate tokens at any possible location. The position of the next token should be given as additional input token or as an additional positional embedding. (\textbf{Bottom}) The greedy order generation uses the any-order model but for each token generates all possible positions and selects only the most likely one (darker yellow) as the next generated token.}
    \label{fig:models}
\end{figure}

We find that the answer is yes. 
As illustrated in Fig.~\ref{fig:inpainting_editing}, each generated image exhibits a specialized generation order—shown using a color gradient from yellow to violet. We also show that our model can preserve the flexible abilities of any order AR and improve generation quality. 
To achieve this, we present Ordered Autoregressive (OAR), we build on recent advancements in AR modeling~\cite{li2024autoregressive, pannatier2024sigma, pang2024randar}. We first train an AR model to learn chains of conditional distributions under arbitrary orders (an any-order model). We then distill knowledge from this model by re-labeling the training samples without changing the underlying structure, selecting orders at each step of autoregression with the highest prediction likelihood (a greedy order selection). Fig.~\ref{fig:models} visually contrasts raster, any-order, and greedy generation. Finally, we finetune the model using these image--order pairs in a self-supervised manner. Through this process, we observe clear improvements in autoregressive image generation quality at the same time keeping the flexible abilities of any-order generation. 

The contributions of our paper are as follows:
\begin{itemize}
    \item We propose a distillation strategy that extracts specialized generation orders from an any-order AR model's own confidence scores and fine-tunes on them, improving generation quality while preserving the model's ability to perform tasks such as inpainting and outpainting without retraining
    \item We provide a theoretical and empirical analysis showing that order specialization acts as capacity reallocation: by redirecting model capacity from fitting $N!$ permutations to a single specialized trajectory per image, generation quality improves without erasing the any-order conditionals that enable multi-task flexibility.
    \item We validate our approach on ImageNet~\cite{deng2009imagenet}, reporting improved performance over any-order and raster generation. 
    In addition, we validate our method on a Fashion dataset~\cite{fashiondataset} and Multimodal CelebA-HQ dataset~\cite{xia2021tedigan}. We further show zero-shot inpainting and outpainting capabilities, and provide ablation studies identifying the key factors of contribution.
\end{itemize}

\section{Related Work}

\noindent\textbf{Autoregressive Image Generation.}
AR models, long dominant in language modeling, have recently matched the performance of diffusion models~\cite{austin2021structured}, renewing interest in AR methods for vision~\cite{yu2024language, sun2024autoregressivemodelbeatsdiffusion}. Early AR image generators~\cite{van2016pixel, van2016conditional, chen2018pixelsnail, parmar2018image, chen2020generativepretraining} produced images pixel by pixel using RNNs~\cite{van2016pixel}, CNNs~\cite{van2016conditional}, and later Transformers~\cite{parmar2018image, chen2020generativepretraining}. Subsequent work adopted patch-wise tokenization with learned visual codebooks, yielding substantial improvements~\cite{van2017neural, esser2021taming, razavi2019generating, ramesh2021zero}. Despite these advances, a large body of work shows that specialzing sequence ordering remains a critical factor influencing AR models’ ability to capture the underlying data distributions differently~\cite{vinyals2016order, ford2018importance, papadopoulos2024arrows}.

\noindent\textbf{Any-Order Generation.}
Any-order AR image generation models~\cite{uria2016neural, yang2019xlnet, hoogeboom2022autoregressive} allow generation under arbitrary token orders. Encoder-based models such as MaskGIT~\cite{chang2022maskgit} and MAGE~\cite{li2023mage} follow a BERT-style masked modeling approach~\cite{devlin2019bert} with iterative decoding driven by masking schedules and confidence scores. The MAR framework~\cite{li2024autoregressive, fan2024fluid} unifies these techniques by interpreting masked generation as next-token prediction under flexible orders. Unlike masked encoder-based models~\cite{chang2022maskgit, li2023mage, li2024autoregressive, fan2024fluid}, our approach is decoder-only, enabling efficient KV-caching.

Interestingly within NLP, any-order AR models have been used to exploit bidirectional context modeling: XLNet~\cite{yang2019xlnet} optimizes the expected likelihood over all permutations, while $\sigma$-GPT~\cite{pannatier2024sigma} dynamically selects per-sample generation orders. Recent work, such as RAR~\cite{yu2024randomized}, trains on randomized orders that gradually anneal toward raster order, leveraging regularization benefits.
We build on this line of work, we argue with order specialization, we can achieve measurable quality gains and at the same time maintain the flexibility of any-order AR image generation, which is crucial for zero shot applications like inpainting and outpainting.

\noindent\textbf{Learning special orders in Autoregressive Models.}
Various methods seek to determine different autoregressive orderings through techniques like bidirectional decoding~\cite{sun2017bidirectional, mehri2018middle} or syntax tree-based decoding~\cite{yamada2001syntax, wang2018tree,welleck2019non}. These approaches, however, rely on heuristics rather than learning orders spacialized for different samples. InDIGO~\cite{gu2019insertion} incorporates insertion operations to enable arbitrary-order generation. Variational Order Inference (VOI)~\cite{li2021discovering} and LO-ARM~\cite{wang2025learningorder} learns specific orders via variational inference, requiring joint training of an order encoder and a generative decoder and relying on high-variance estimators (e.g., REINFORCE) because the discrete order sampling breaks differentiability.
In contrast, our method learns specialized orders through a simple self-distillation pipeline: we decouple order discovery from generation by training a general any-order model, then greedily extract specialized orders based on the model’s own confidence scores, and finally fine-tune the model on these specialized orders. This approach is simple, efficient, and does not require complex joint training or high-variance estimators.
\section{Perspective and Motivations}
The generation order in an AR model determines both what the model can do and how well it does it~\cite{li2024autoregressive}. While later on the experiment section presents a comprehensive analysis of the results, this section provides a deeper understanding of the direct and indirect effects of orders for an AR model and our formulation is detailed in the following section.

\subsection{Joint and Conditional chain Equivalence}
\label{sec:ar-aoar}

Let a sequence $\mathbf{x} = [x_1, x_2, \ldots, x_N]$ be an i.i.d.\ sample of discrete tokens.  
In generative modeling, the fundamental objective is to learn the underlying data distribution $p(\mathbf{x})$, which encodes the structure and variability of the data. Once this distribution is learned, the model can generate new, realistic samples that preserve the essential characteristics of the true data.

To make likelihood computation tractable, AR models factorize the joint distribution using the chain rule,
$p_\theta(\mathbf{x}) = \prod_{i=1}^{N} p_\theta(x_{l_i} \mid \mathbf{x}_{\mathbf{l}_{<i}})$,
where $\mathbf{l}$ is a permutation of $\{1,\ldots,N\}$, and $l_i$ and $\mathbf{l}_{<i}$ denote the $i$-th position and the first $i-1$ positions of the chosen ordering, respectively.  
Sampling from the model then proceeds sequentially according to this ordering.  
If the AR model perfectly captured the true joint distribution, the factorization order would be inconsequential, as the product of conditional probabilities would exactly match the true joint probability.

When training a model iteratively on a single fixed ordering (e.g., raster-scan), the model learns to predict tokens in that specific sequence and can potentially introduce an inductive bias that may not align well with the underlying data structure, especially for data with no inherent or natural structure~\cite{vinyals2016order}. Thus in practice, the choice of ordering can significantly impact the model's ability to learn and generalize effectively for a model trained on a sufficiently large dataset.

A key factor in optimizing an AR model $p_\theta$ is therefore the order in which sequence elements are arranged during training. 
Any-Order AR models generalize standard AR models by enabling generation under \emph{any} prescribed ordering of the input sequence.  
To this end, we train the model using random orderings drawn uniformly from the permutation set
$\mathcal{P}_L = \{ \mathbf{l} \mid \mathbf{l} \text{ is a permutation of } \{1,\ldots,N\} \}$, with $|\mathcal{P}_L| = N!$.
Let $p(\mathbf{l})$ denote the uniform distribution over $\mathcal{P}_L$.  
The parameters $\theta$ are learned by maximizing the expected log-likelihood over these orderings:
$\mathbb{E}_{p(\mathbf{l})}\!\left[ \log p_\theta(\mathbf{x} \mid \mathbf{l}) \right]$.

This objective can be interpreted as a variational lower bound on the marginal log-likelihood of a latent-variable model~\cite{uria2014deep}.  
Treating each permutation $\mathbf{l}$ as a latent variable associated with $\mathbf{x}$, we have:
\begin{equation*}
\log p_\theta(\mathbf{x})
= \log \sum_{\mathbf{l} \in \mathcal{P}_L} p(\mathbf{l})\, p_\theta(\mathbf{x} \mid \mathbf{l})
\;\ge\;
\mathbb{E}_{p(\mathbf{l})}\!\left[ \log p_\theta(\mathbf{x} \mid \mathbf{l}) \right].
\end{equation*}

In practice, this expectation is approximated by stochastically sampling a permutation for each training instance.

\subsection{Effect of Generation Order on Likelihood Optimization}
\label{sec:order-gen}

All possible chains of conditional distributions can be viewed as paths in a tree, where each level corresponds to a generation step and each branch represents a possible outcome for the current variable. During training, however, only a single path (i.e., a single ordering) is followed for each sample. Thus, learning effectively amounts to identifying which paths through this tree are most beneficial for modeling the data distribution.

At each level $i$ of this tree, the model $p_\theta$ must learn a conditional likelihood $p(\cdot \mid \mathbf{x}_{<i})$ so that, across $N$ levels $(x_0, x_1, \ldots, x_N)$, it approximates the true data distribution $p(\mathbf{x})$. We argue that selecting high-probability locations for the next token at each level is crucial for two key reasons:  
(1) \emph{compounded probabilities}, and  
(2) \emph{the lack of a recovery mechanism} in AR generation.

\noindent\textbf{Compounded probabilities.}  
Because likelihoods at level $i$ directly depend on the choice made at level $i{-}1$, a low-probability choice early in the sequence may force the model into an unlikely or suboptimal region of the tree. This leads to poor likelihood estimates at deeper levels, therefore, can be effecitvely pruned out as an effective order and can potentially saturate model capacity. 

\noindent\textbf{No recovery mechanism.}  
Due to the sequential nature of autoregression, early decisions strongly constrain future ones. The standard generation procedure samples  $c_i \sim p_\theta(c_i \mid c_{<i})$, producing one token at a time based on previous outputs.  
If an early sampled token has low likelihood under the data distribution, the model is pushed into regions with little quality training support, often leading to cascading errors.  
Conversely, generation orders that begin with high-likelihood tokens keep the model within high-density regions of the data, reducing the chance of drifting into implausible generations and improving robustness.

\section{Our Method: Ordered AR Self-Distillation}
\subsection{Problem Interpretation: Order Specialization as Capacity Reallocation}
Any order training allows the model to be flexible to zero shot applications by learning conditionals under $N!$ permutations, where $N$ is the number of patches. This property of the model forces its finite capacity to be distributed across all these $N!$ permutations. By extracting a specialized order, we can reallocate the capacity previously spent on learning the permutation space to instead refine generation quality under a specialized and data-driven order. This significantly simplifies the learning problem and leads to improved AR image generation.

We therefore seek to learn a specialized order that maximizes the likelihood of the training data, while still retaining the flexibility of any-order generation. To make this problem tractable, we first train an any order AR model that performs well under \emph{any} ordering permutation, as described in section~\ref{sec:ar-aoar}. We then extract a specialized order for each training sample by maximizing the likelihood under the learned model, as described in section~\ref{sec:order-gen}. Finally, we fine-tune the model with these image--order pairs, allowing it to repurpose the capacity previously spent on learning the permutation space to instead refine generation quality under a specialized and data-driven order. This significantly simplifies the learning problem and leads to improved AR image generation.

\subsection{Training Any-Order AR Model}

As in standard autoregressive pipelines~\cite{sun2024autoregressivemodelbeatsdiffusion, esser2021taming}, an image and its conditioning signal (e.g., a class label or text prompt) are first converted into a sequence of discrete tokens using a standard VAE~\cite{esser2021taming,sun2024autoregressivemodelbeatsdiffusion} encoder. Our method is orthogonal to the choice of encoder and operates directly on these discrete representations.

To enable generation under arbitrary orders, the model must be informed of \emph{which} token will be generated next. We explore two established mechanisms for providing this “next-token” positional information:

\begin{itemize}
    \item {RandAR-style position tokens}~\cite{pang2024randar}:  
    Each content token $x_i$ is paired with an auxiliary token $l_i$ that explicitly encodes the position of the next token to be predicted. The model therefore receives $(x_i, l_i)$ as a combined input at every step.

    \item {$\sigma$-GPT-style double positional encoding}~\cite{pannatier2024sigma}:  
    Instead of introducing a separate token, each content token is embedded using two positional encodings—one for its actual spatial location and one for the spatial location of the next token to be generated.
\end{itemize}

Both approaches provide the AR transformer with knowledge of the upcoming position, enabling it to model the conditional distribution for \emph{any} permutation of the image patches. This yields our Any-Order AR (AO-AR) model, which forms the foundation for the subsequent phases of the method. We use standard cross-entropy next token prediction as the training objective.

\subsection{Specialized Order Extraction}
\label{sec:phase2}

Once the any-order model has been trained, we use it to extract a specialized generation order for every training image. The goal of this phase is to use the model’s own predictive structure to oobtain an ordering that prioritizes high-confidence positions, mitigating the compounding and recovery issues discussed in Section~\ref{sec:order-gen}.

We avoid the use of beam search method in NLP~\cite{pannatier2024sigma} to explore multiple candidate orders, due to its substantial computational and memory overhead, particularly prohibitive when $N$ is large. Instead, we adopt an efficient greedy extraction procedure that relies solely on the model’s confidence scores.

Formally, the specialized order $\mathbf{l}^*$ is obtained via teacher forcing. At each step $i$ (from $1$ to $N$), the model receives the ground-truth prefix $(\mathbf{x}_{<i}, \mathbf{l}_{<i})$ and outputs a probability distribution over all remaining patch locations $L_i$. For each candidate location $l \in L_i$, the model produces a distribution over the image-token vocabulary $V$. We assign to each location the maximum attainable likelihood over $V$, and select the next position as:

\begin{equation}
\label{eq:greedy}
l^*_i
= \arg\max_{l \in L_i}
\left[
\max_{v \in V}
p_\theta(v \mid l, \mathbf{x}_{<i}, \mathbf{l}_{<i})
\right].
\end{equation}

This greedy rule prioritizes locations that the model is most confident about predicting at each step. Repeating this procedure for all $N$ steps yields a complete content-aware ordering $\mathbf{l}^*$ for the image. Collectively, these pseudo-labeled pairs $(\mathbf{x}, \mathbf{l}^*)$ constitute a self-supervised curriculum that guides the final self-distillation.

\subsection{Self-Distillation}

The final stage consists of fine-tuning the any-order model using the image--order pairs extracted. While the AO-AR model learns to handle \emph{all} possible orderings, in this phase we focus on repurposing the model capacity to specialize its predictive performance on the distilled orders $\mathbf{l}^*$.


To avoid catastrophic forgetting of this general knowledge and keep the flexibility of AO-AR models, we fine-tune using a reduced learning rate, ensuring that the model adapts to the specialized orders while preserving the robustness and flexibility acquired earlier. In addition, we incorporate an order-annealing schedule similar to RAR~\cite{yu2024randomized}.  
This schedule gradually transitions from training on mixed or partially randomized orders to training purely on the specialized orders, which empirically stabilizes model performance and further improves generation quality.

Together, these choices allow the model to retain the strengths of the any-order formulation while specializing toward a single, specialized generation path. Exploring iterative refinement, alternating between order estimation and fine-tuning remains an interesting direction for future work.

\subsection{Inference with OAR}
During inference, we apply the same greedy technique described in ~\eqref{eq:greedy}. However, instead of using ground-truth tokens, the context is composed of the tokens generated in the previous steps. With this approach, we ensure that the model generates tokens from the easiest to the most difficult, which improves image quality by reducing the risk of error accumulation.

While the computational complexity during training remains comparable to that of raster-scan AR models, our inference procedure results in a higher computation cost as it requires $\frac{N(N+1)}{2}$ forward pass to generate an image, where N denotes the number of patches in an image. However, we employ the following optimizations to keep the time complexity low. In practice, we find that the inference time of our method is comparable to that of standard AR models. We attribute this to the power of modern GPUs, which can efficiently parallelize the computations required for evaluating multiple candidate locations at each step.

\noindent\textbf{Parallel Querying.} At each generation step i, we evaluate all $N-1+i$ possible next locations in parallel, in a single forward pass. For the RandAR approach, each input consists of the previously generated tokens interleaved with their corresponding position tokens, followed by the position tokens for all possible next locations (queries). To ensure predictions are independent, the attention mask is modified at each step to prevent queries from attending to each other.

\noindent\textbf{Optimized KV-Caching.} For the RandAR approach, we only populate the KV-Cache with the interleaved image and position tokens corresponding to the previously selected locations. For the $\sigma$-GPT approach, we adapt the KV-caching mechanism \cite{ramesh2021zero} to speed up inference and provide a pseudo-code of our implementation in the Appendix.

Furthermore, our framework still allows parallel decoding. By replacing the $argmax$ in ~\eqref{eq:greedy} with a top-k selection, our model can generate $k$ tokens in parallel. Our method is compatible with standard LLM inference optimizations (batching, mixed precision, flash attention); details in Appendix.

\section{Experiments on ImageNet}
\label{sec:imagenet_exp}
We conduct our primary experiments on the class-conditional ImageNet~$256\times256$ benchmark ~\cite{deng2009imagenet}. 
This section first studies how different generation orders—Raster, Random, and our Greedy (OAR) order—affects the performance of a RandAR-XL backbone under various fine-tuning regimes. We analyze the impact of order choice during sampling, the role of specialized order fine-tuning, and the effect of order annealing.

In the later part of this section, we further compare our method against state-of-the-art autoregressive and diffusion-based generative models to contextualize the performance gains achieved by our specialized ordering strategy. More details pertaining to our training and evaluation can be found in Appendix.

\subsection{Fine-tuning and Sampling orders}
In Tab.~\ref{table:imagenet_finetune}, we report the effect of different generation orders when fine-tuning a RandAR-XL model~\cite{pang2024randar}. We compare three fine-tuning strategies (\emph{None}, \emph{Raster}, and \emph{Greedy}) and evaluate each model under Raster, Greedy, and Random sampling. Without fine-tuning, random sampling performs best (FID~2.39), while Greedy (FID~3.64) and especially Raster (FID~12.31) perform worse. Although random sampling should theoretically cover all orderings, in practice it induces sampling statistics that differ substantially from structured patterns such as raster or greedy, which explains the performance drop for those two.
\begin{table}[ht!]
\centering
\caption{Comparisons of different order generation on class-conditional ImageNet 256×256 benchmark. For a fair evaluation, all tests are based on RandAR-XL with 256 step generation. The inpainting metrics were calculated on the ImageNet validation set (masking 25\% to 75\%). 
}
\label{table:imagenet_finetune}

\vspace{-2mm}
\resizebox{.7\linewidth}{!}{
\begin{tabular}{c|c|c|c|c|c|c}
\toprule
\textbf{Fine-Tuning} & \textbf{Annealing} & \textbf{Generation} & \textbf{FID↓} & \textbf{IS↑} & \textbf{Prec.↑} & \textbf{Recall↑} \\
\midrule
None & \ding{56} & Raster & 12.31 & 127.55 & 0.59 & 0.70 \\
 & \ding{56} & Greedy & 3.64 & 220.48& 0.61 & 0.74 \\
 & \ding{56} & Random & 2.39 & 253.91 & 0.80 & 0.60 \\

\midrule
Raster & \ding{56} & Random  & 6.57 & 160.18 & 0.70 & 0.65 \\
& \ding{56} & Raster & 2.56 & 236.15 & 0.79 & 0.60 \\
& \ding{51} & Raster & 2.34 & 242.54 & 0.79 & 0.61 \\

\midrule
Greedy & \ding{56} & Random & 2.44 & 281.92 & 0.80 & 0.59 \\
 & \ding{56} & Greedy & 2.22 & 256.77 & 0.80 & 0.60 \\
& \ding{51} & Greedy & 2.17 & 258.37 & 0.80 & 0.60 \\
\midrule\midrule
\multicolumn{7}{c}{\small \emph{Inpainting (mask 25\%--75\%)}} \\
\midrule
None & \ding{56} & Random & 2.14 & 280.09 & 0.79 & 0.60 \\
Greedy    & \ding{51} & Greedy & 2.10 & 286.39 & 0.85 & 0.63 \\
\bottomrule
\end{tabular}
}
\end{table}
Fine-tuning with raster order yields modest gains over random sampling. This gap suggests that simply reducing from $N!$ to one fixed order is not sufficient---the choice of order matters. Specialized orders allow the model to reallocate capacity from fitting all $N!$ conditional chains to refining a single, confidence-driven generation path. Unlike RAR~\cite{yu2024randomized}, which trains from scratch with order annealing and thus conflates the benefits of order specialization with training dynamics, our setup isolates the effect of order choice, confirming that the quality gain stems from the specialized order itself.
We also evaluate the inpainting performance of our approach compared to the any-order model with masking ranging from $25\%$ to $75\%$ of the image size. Our approach produces slightly better images (OAR 2.10 FID vs. Any-order 2.14 FID) due to our specialized order. 

\subsection{Comparison with State-of-the-art}

\begin{table*}[ht!]
\centering
\caption{Comparisons of models below 1B parameters on class-conditional ImageNet $256\times 256$ benchmark from \cite{pang2024randar}. Metrics are Fréchet inception distance (FID), inception score (IS), precision and recall. ``↓'' or ``↑'' indicate lower or higher values are better. ``-re'' means using rejection sampling. $^*$ represents training at 384x384 resolution, and resized to $256\times 256$ for evaluation.  
}\vspace{-2mm}
\label{table:main_results}
\resizebox{0.9\linewidth}{!}{
\begin{tabular}{c|l|c|c|c|c|c|c}
\toprule
\textbf{Type} & \textbf{Model} & \textbf{\#Para.} & \textbf{FID↓} & \textbf{IS↑} & \textbf{Precision↑} & \textbf{Recall↑}  & \textbf{Steps}\\
\midrule
GAN & BigGAN \cite{brock2018large} & 112M & 6.95 & 224.5 & 0.89 & 0.38 & 1\\
& GigaGAN \cite{kang2023scaling} & 569M & 3.45 & 225.5 & 0.84 & 0.61 & 1 \\
& StyleGan-XL \cite{sauer2022stylegan} & 166M & 2.30 & 265.1 & 0.78 & 0.53 & 1 \\
\midrule
Diffusion & ADM \cite{dhariwal2021diffusion} & 554M & 4.59 & 186.70 & 0.82 & 0523 & 250 \\
& LDM-4 \cite{rombach2022high} & 400M & 3.60 & 247.7 & -- & -- & 250\\
& DiT-XL \cite{peebles2023scalable} & 675M & 2.27 & 278.2 & 0.83 & 0.57 & 250\\
& SiT-XL \cite{ma2024sit} & 675M & 2.06 & 270.3 & 0.82 & 0.59 & 250 \\
\midrule
Bi-directional AR & MaskGIT-re \cite{chang2022maskgit} & 227M & 4.02 & 355.6 & -- & -- & 8 \\
& MAR-L\cite{li2024autoregressive} & 479M & 1.98 & 290.3 & -- & -- & 64 \\
& MAR-H\cite{li2024autoregressive} & 943M & 1.55 & 303.7 & 0.81 & 0.62 & 256 \\
& TiTok-S-128~\cite{yu2024image} & 287M & 1.97 & 281.8 & -- & -- & 64 \\
\midrule
Causal AR& VAR \cite{tian2024visual} & 600M & 2.57 & 302.6 & 0.83 & 0.56 & 10 \\
& SAR-XL \cite{liu2024customize} & 893M & 2.76 & 273.8 & 0.84 & 0.55 & 256 \\
& RAR-B \cite{yu2024randomized} & 261M & 1.95 & 290.5 & 0.82 & 0.58 & 256 \\
& RAR-L \cite{yu2024randomized} & 461M & 1.70 & 299.5 & 0.81 & 0.60 & 256 \\
& RAR-XL \cite{yu2024randomized} & 955M & 1.50 & 306.9 & 0.80 & 0.62 & 256 \\
& LlamaGen-L\cite{sun2024autoregressivemodelbeatsdiffusion}  & 343M & 3.07 & 256.06 & 0.83 & 0.52 & 256 \\
& LlamaGen-XL$^*$\cite{sun2024autoregressivemodelbeatsdiffusion} & 775M & 2.62 & 244.08 & 0.80 & 0.57 & 576 \\
 & RandAR-L~\cite{pang2024randar} & 343M & 2.65 & 249
 & 0.82 & 0.56 & 256 \\
& RandAR-XL~\cite{pang2024randar} & 775M & 2.39 & 253.91 & 0.80 & 0.60 & 256 \\ 
& OAR (Ours w/o Annealing) & 775M & 2.22 & 256.37& 0.80& 0.60& 256\\
& OAR (Ours w/ Annealing) & 775M & 2.17 & 258.37& 0.80& 0.60& 256\\

\bottomrule
\end{tabular}
}
\end{table*}

Tab.~\ref{table:main_results} summarizes comparisons against sub-1B parameter models on ImageNet~$256\times 256$. OAR provides consistent improvements over the RandAR baseline: fine-tuning with our specialized order reduces the FID from 2.39 (RandAR-XL, 256 steps) to 2.22, and further to 2.17 with annealing. This indicates that adapting the model to a learned generation path yields measurable gains while retaining the flexibility of the any-order formulation (see Table~\ref{table:imagenet_finetune}). Within the family of causal AR models, OAR is competitive at similar model scales. While RAR-XL achieves a lower FID, it converges to a fixed raster order, forfeiting the any-order conditionals needed for tasks like inpainting. OAR provides a better trade-off: competitive generation quality with preserved multi-task capabilities.

OAR operates on a lightweight fine-tuning regime on a pretrained RandAR model and achieves stronger performance than LlamaGen-XL (2.62 FID) under the same parameter budget, which also serves as the core backbone for all our experiments. Finally, because OAR preserves RandAR's multi-patch decoding capability, it supports accelerated sampling by predicting the top-$k$ most confident locations per step. Using 88 decoding steps, OAR attains a FID of 2.51, offering an efficient speed–quality trade-off without architectural changes.

\subsection{Human Evaluation}

While FID captures distribution-level statistics, it can fail to reflect fine-grained visual artifacts. To assess the perceptual impact of our ordering strategy, we conducted a forced-choice preference study.
\begin{table}[ht!]
\centering
\caption{\textbf{Human Preference Study (OAR vs. RandAR-XL).} Our model achieves a substantial \textbf{64.33\%} average preference rate across 21 ImageNet classes. Notably, OAR is preferred in 16 out of 21 comparisons, with 57\% of classes (12/21) showing a "strong preference" (>65\% win rate).}
\label{tab:human_results_summary}
\small
\begin{tabular}{l|c|c|c}
\toprule
\textbf{Metric} & \textbf{Win Rate (Avg.)} & \textbf{Preferred Classes} & \textbf{Strong Wins ($>65\%$)} \\
\midrule
OAR (Ours) vs. Baseline & \textbf{64.33\%} & \textbf{16 / 21} & \textbf{12 / 21} \\
\bottomrule
\end{tabular}
\end{table}
\paragraph{Study Design.} We sampled 21 ImageNet classes and generated image pairs (OAR vs. RandAR-XL baseline). Each pair was evaluated by 51 independent participants (1,071 total judgments) with randomized presentation to eliminate bias.

\paragraph{Results.} As summarized in Table~\ref{tab:human_results_summary}, OAR significantly outperforms the baseline with an average preference rate of \textbf{64.33\%}. Notably, our model maintains the majority preference in \textbf{76.2\%} of the tested categories (16 out of 21). Crucially, these results demonstrate a strong alignment between automated metrics and human perception; the gains in distributional fidelity (FID) translate directly into images that are subjectively more coherent and visually appealing. Detailed per-class results and qualitative comparisons are provided in Appendix.

\section{Experiments on Text-to-Image}
\label{sec:t2i_exp}
\subsection{Datasets}
In this section, we empirically evaluate the performance of our model under text-to-image setting and attempt to analyze our findings. We evaluate Fréchet inception distance (FID), Inception Score (IS), and Kernel Inception Distance (KID)~\cite{kynkaanniemi2023the} on the dataset. In addition, we report also the average distance between two subsequent generated patches (d). The corresponding ablations to this dataset can be found in the Appendix. 
Evaluations are conducted on the Fashion Product dataset~\cite{fashiondataset} and additionally evaluated on Multimodal CelebA-HQ dataset~\cite{xia2021tedigan}. The fashion dataset contains 44,400 images of fashion products and their corresponding captions. The CelebA datasset contains $30,000$ images and corresponding attributes as captions. For both the datasets, we use $90\%$ of the data for training and the other $10\%$ for testing purposes. Each image has a white background and an object at the center for the Fashion Product Dataset whereas three visual descriptions are provided for each image, with more or less details for the Multimodal CELEBA-HQ dataset.



\subsection{Implementation Details}
In our experiments, we train a single decoder-only transformer. We use pretrained VQGAN~\cite{esser2021taming} encoder to encode the images, decoder, and codebook of 1024 tokens. The encoder uses a patch size of $16\times 16$ to encode the images. 

The same configuration is used to train all models: for each image, we randomly chose a caption in the runtime among the multiple captions provided by the dataset. The text sequence is truncated to 256 tokens when longer, padded with special tokens when shorter, and concatenated to the image tokens sequence. The transformer uses 768 embeddings in dimensions, a depth 6, and 14 heads of 64 dimensions each. We use learnable positional embeddings that are additively factorized for parameter efficiency~\cite{ho2020axial}. We use layer normalization \cite{ba2016layernormalization} and apply dropout \cite{dropout} to linear projections in both feed-forward networks and self-attention layers, with a probability $p = 0.2$. We apply randomly resized crop augmentation of the data for the images. We weight the loss computed on image tokens by a factor of 7 compared to the loss for text tokens, following~\cite{ramesh2021zero}. We use the AdamW optimizer~\cite{loshchilov2018decoupled} with a learning rate of $3 \times 10^{-4}$ that we reduce by a factor of 0.8 following the ReduceLROnPLateau schedule. Models are trained for 300 epochs with a batch size 16 on each machine.


\begin{table}[ht!]

    \centering
    \caption{Generation with different orders. Our ordered generation (OAR) improves over the standard raster-scan order and random order generation similar to~\cite{li2024autoregressive}. FID is the Fréchet inception distance, IS is the Inception Score and KID Kernel Inception Distance. D denotes the average distance between subsequently generated patches.}
    \begin{adjustbox}{max width=\linewidth}
    \begin{tabular}{l|cc|ccc|c}
    \toprule
       \textbf{Method} & \textbf{Train} & \textbf{Generation} & \textbf{FID} $(\downarrow)$ & \textbf{IS} $(\uparrow)$  & \textbf{KID} $(\downarrow)$ & \textbf{d}\\
   \midrule
    Raster & Raster & Raster & 4.58 & 1.106 & 0.0031 & 1.83\\
    Random-Raster & Random & Raster & 4.38 & 1.102  & 0.0031 & 1.83 \\
    Random & Random & Random & 4.07 & 1.103  & 0.0028 & 8.19\\
    FT Ordered (Ours) & Ordered & Ordered & 2.56 & 1.111 & 0.0015 & 3.99\\
     \bottomrule
    \end{tabular}
    \label{tab:gen_rslt}
    \end{adjustbox}
\end{table}

\subsection{Fashion Products}
\label{sec:image_gen_metrics_fashion}

\begin{figure*}[ht!]
    \centering
    \includegraphics[width=0.99\linewidth,keepaspectratio]{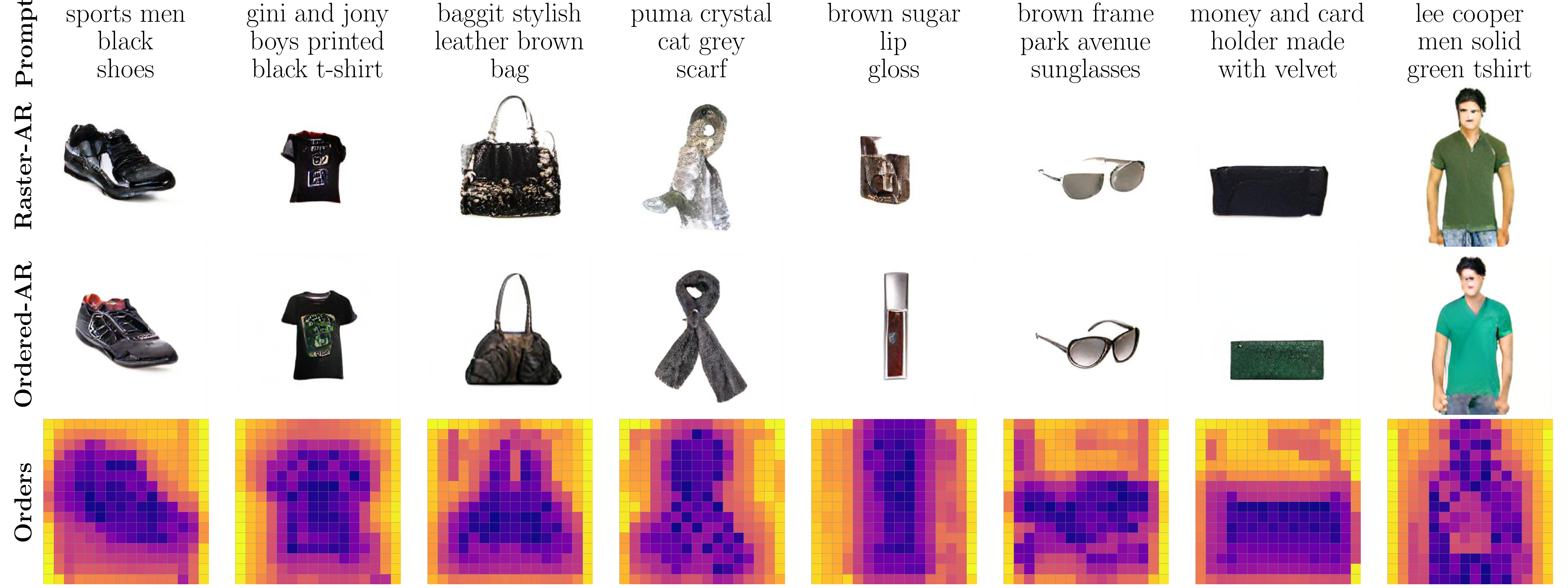}
    \caption{Examples of generation on the Fashion Products dataset. \textbf{(Top)} Generated images with raster AR mode. \textbf{(Middle)} Generated images with OAR model.
    \textbf{(Bottom)} Generation order, from yellow to violet. From these images, we see that our approach finds an order highly correlated with the image content, often resulting in better image quality.}
    \label{fig:samples_fashion}
\end{figure*}

Table~\ref{tab:gen_rslt} presents results for different training and generation orders on the Fashion Products datasets. 
As already shown in other works \cite{li2024autoregressive, fan2024fluid, yu2024randomized}, we observe that the \emph{Raster scan} order yeilds inferior results and that we can get better generation performance by training with \textit{Random} orders. By generating with specialized orders the most likely patches first, our model reduces possible drifting, consistently improving the generation. This is also reflected through our ImageNet expeirments where we observe similar performance.
Finally, the best performance is obtained when our model is also \textit{Fine-tuned} with specialized orders. notably this is also consistent with our other results. In Figure \ref{fig:samples_fashion}, we present examples of images generated by our method and the corresponding generation orders and compare them with raster-scan generation. 

\subsection{Multimodal CelebA-HQ dataset}

Additionally, we experiment on the Multi-Modal CelebA-HQ dataset using a random $90:10$ train-test split. This dataset contains $30,000$ celebrity face images with diverse facial attributes such as eyebrow shape, iris color, makeup styles, and hair types, among others. Unlike the Fashion Product dataset, CelebA-HQ features varied backgrounds; however, the primary focus remains on facial features and attributes. The primary target is to generate a face based on a caption that consists of facial attributes
    
\begin{table}[ht!]
    \centering
    \caption{Generation with Different Orders on the Multi-Modal CelebA-HQ Dataset. Our ordered generation improves over the standard raster-scan order.}
    \begin{tabular}{c|c}
    \toprule
    \textbf{Model} & \textbf{FID} $(\downarrow)$\\
    \midrule
    Raster-Scan AR& 1.94 \\
    Ordered AR & 1.52 \\
    Fine-Tuned Ordered AR& 1.41 \\
    \bottomrule
    \end{tabular}
    \label{tab:celeba}
\end{table}
Consistent with our other experiments, OAR improves over both raster and any-order baselines (Tab.~\ref{tab:celeba}). The gain from raster (1.94) to fine-tuned ordered (1.41) is substantial, confirming that the self-distillation pipeline generalizes across datasets and conditioning modalities.

The specialized orders exhibit a semantically structured progression: the model first generates high-confidence regions such as cheeks and chin, then progresses to fine-grained features like eyes, lips, and hair, and completes the background last. Notably, the model tends to complete one semantic region before moving to the next, effectively decomposing the generation into coherent sub-tasks. This aligns with our capacity reallocation argument---specialized orders simplify each conditional by ensuring semantically coherent context at every step. Qualitative examples and additional experiments with randomized backgrounds are provided in the Appendix.






\section{Conclusion}

We presented Ordered Autoregressive (OAR) image generation, where we extract specialized generation orders from an any-order AR model through self-distillation. By fine-tuning on orders derived from the model's own confidence scores, OAR improves generation quality while retaining the flexibility of any-order models—supporting zero-shot inpainting, outpainting, and editing without retraining. Experiments on ImageNet, Fashion Products, and CelebA-HQ confirm consistent gains over both raster and any-order baselines, corroborated by human evaluation.

An interesting direction for future work is iterative refinement, where order extraction and fine-tuning alternate to progressively improve order quality. We also see potential in learning an order predictor that avoids greedy extraction at inference, which would reduce computational cost and open the door to higher resolutions and larger models.

\bibliographystyle{iclr2026_conference}
\bibliography{Paper/citations}

\appendix
\onecolumn
{\centering
\Large\textbf{Distilling Specialized Orders for Visual Generation}\\[0.3em]
\large Supplementary Material\\[1.5em]}

{\centering\large\textbf{Table of Contents}\\[0.5em]}
\begin{itemize}[leftmargin=1.5em, itemsep=0pt, parsep=2pt]
  \item \hyperlink{secapprelated}{A\quad Additional Related Work}
  \begin{itemize}[leftmargin=1.5em, itemsep=0pt, parsep=1pt]
    \item \hyperlink{secapporders}{A.1\quad Orders in Autoregressive Models}
    \item \hyperlink{secapppos}{A.2\quad Positional Encodings and Orders}
  \end{itemize}
  \item \hyperlink{secappc2i}{B\quad Class-to-Image}
  \begin{itemize}[leftmargin=1.5em, itemsep=0pt, parsep=1pt]
    \item \hyperlink{subsecparallelquery}{B.1\quad Parallel Query Evaluations via Masking}
    \item \hyperlink{secappkvcache}{B.2\quad Optimized KV Caching via Overwriting}
    \item \hyperlink{secappinsetup}{B.3\quad ImageNet Experimental Setup}
    \item \hyperlink{secappinqual}{B.4\quad Qualitative Samples}
    \item \hyperlink{secapphuman}{B.5\quad Human Evaluation Protocol, Study Design, and Results}
  \end{itemize}
  \item \hyperlink{secappt2i}{C\quad Text-to-Image}
  \begin{itemize}[leftmargin=1.5em, itemsep=0pt, parsep=1pt]
    \item \hyperlink{secappt2igen}{C.1\quad Generation Process}
    \item \hyperlink{secappvinfo}{C.2\quad V-information}
    \item \hyperlink{secappgenapproaches}{C.3\quad Different Generation Approaches}
    \item \hyperlink{secappt2ikv}{C.4\quad KV Caching}
    \item \hyperlink{secappbg}{C.5\quad Effect of Background}
    \item \hyperlink{secappdist}{C.6\quad Distance Regularization}
    \item \hyperlink{secappdecode}{C.7\quad Decoding Strategies}
    \item \hyperlink{secappdistill}{C.8\quad Impact of Self-Distillation}
    \item \hyperlink{secappceleba}{C.9\quad Qualitative Samples on the Multimodal CelebA-HQ}
  \end{itemize}
\end{itemize}

\normalsize




\clearpage

\section{Additional related work}
\hypertarget{secapprelated}{}\label{sec:app_related}
\subsection{Orders in Autoregressive Models}
\hypertarget{secapporders}{}\label{subsec:app_orders}
Research has shown that the order in which we organize input and/or output data in the sequence-to-sequence framework matters significantly when learning an underlying model~\cite{vinyals2016order}. Such insights have led to the development of Any-Order Autoregressive (AO-AR) Models, in which the autoregressive model does not impose a fixed order of generation. Some methods~\cite{germain2015made, uria2016neural} train a single deep feed-forward neural network that can assign a conditional distribution to any variable given any subset of the others. During inference, they create an ensemble of models by sampling a set of orderings, computing the likelihood under each order, and averaging, thus baking an “orderless” inductive bias into the model. Further work~\cite{shih2022training} refines the space of orderings used during training to avoid the redundancy present in previous probabilistic models by training on a smaller set of univariate conditionals that still maintain support for efficient arbitrary conditional inference. In the continuation of AO-ARM approaches, more research has been conducted.

 In Natural Language Processing (NLP) specifically,  motivated by enabling AR language models to learn bidirectional contexts, XLNet~\cite{yang2019xlnet} maximizes the expected log-likelihood over all permutations of the factorization order by relying on a proper attention mechanism in Transformers. $\sigma$-GPT~\cite{pannatier2024sigma} learns a model that can generate sequences of text tokens in any order and modulates the generation order on the fly per-sample during inference in a fashion already explored for Neural Architecture Search~\cite{guo2020single}.

\subsection{Positional Encodings and Orders}
\hypertarget{secapppos}{}\label{subsec:app_pos}
Since the self-attention mechanism in transformers is inherently permutation-invariant, it does not account for token order. It requires additional overhead to encode positional information~\cite{vaswani2017attention, position_transformer}. Early methods introduce absolute positional embeddings, encoded through sinusoidal functions  or learnable embeddings, to differentiate tokens based on their positions rather than just their content~\cite{dosovitskiy2020image, vaswani2017attention}. However, as the number of tokens grows, absolute positional embeddings become less efficient. To address this limitation, Shaw \etal~\cite{shaw2018self} introduces relative position biases directly into the attention matrix within the layers of self-attention, enhancing spatial awareness and scalability. Although both absolute- and relative position embeddings are effective in fixed-token-size scenarios, they struggle with adapting to variable token sizes and require flexibility and extrapolation capabilities. To overcome these challenges, Su \etal~\cite{su2024roformer} proposes an approach that encodes absolute positions using a rotation matrix and integrates explicit relative positional dependencies into the self-attention formulation. This combined method enhances the adaptability to different resolutions, enabling the model to generalize better across varying spatial contexts.

In our work, following the same motivation as~\cite{gu2019insertion}, we leverage relative positions for our text-to-image setting to alleviate the challenges of predicting absolute positions. Our method introduces two key novelties: First, we extend the relative positional encoding to handle 2D inputs effectively. Second, we employ a dual positional encoding scheme—preserving the positional encoding of the current patch as absolute, while encoding the position of the next patch relative to the current one. Notably, our proposed relative positional embedding differs from prior methods, where relative positions are directly embedded within the transformer architecture itself~\cite{yang2019xlnet}. This positional encoding strategy is used throughout OAR's any-order AR backbone for the text-to-image setting.

\section{Class-to-Image}
\hypertarget{secappc2i}{}\label{sec:app_c2i}

For class conditional training on ImageNet~\cite{deng2009imagenet}, we use the architecture from RandAR~\cite{pang2024randar} which is more optimized for the large-scale training . Specifically, we use the RandAR-XL model available through Hugging Face.

\subsection{Parallel Query Evaluations via Masking}
\hypertarget{subsecparallelquery}{}\label{subsec:parallel_query}

To select the next token location, we evaluate $K$ candidate positions in parallel. Let $t$ be the number of previously generated (committed) tokens. In a standard autoregressive setup, appending $K$ queries would cause the $i$-th query to attend to the $j$-th query (where $j < i$), creating a dependency between the candidates that should be evaluated independently.

To batch these evaluations into a single forward pass, we modify the causal attention mask $\mathbf{A}$. For a sequence containing $t$ history tokens and $K$ query tokens, the mask $\mathbf{A} \in \{0,1\}^{(t+K) \times (t+K)}$ is constructed as follows:

\begin{equation}
    \mathbf{A}_{i,j} = 
    \begin{cases} 
    1 & \text{if } i \le t \text{ and } j \le i \quad \text{(Causal History)} \\
    1 & \text{if } i > t \text{ and } j \le t \quad \text{(Queries see History)} \\
    1 & \text{if } i = j \quad \quad \quad \quad \quad \quad \;\; \text{(Self-Attention)} \\
    0 & \text{otherwise}
    \end{cases}
\end{equation}

This structure ensures that while all queries can attend to the full history ($1 \dots t$), the block representing query to query interactions ($i,j > t$) is strictly diagonal. This makes the distribution for each query independent of previous queries. Figure~\ref{fig:attn_mask} illustrates this mask for a scenario with 7 tokens ($t=4$ history tokens and $K=3$ parallel queries).

\begin{figure}[h]
    \centering
    \begin{tikzpicture}[scale=0.5]
        \draw[step=1cm,gray!30,very thin] (0,0) grid (7,7);
        
        \fill[gray!80] (0,6) rectangle (1,7); 
        \fill[gray!80] (0,5) rectangle (1,6); \fill[gray!80] (1,5) rectangle (2,6);
        \fill[gray!80] (0,4) rectangle (1,5); \fill[gray!80] (1,4) rectangle (2,5); \fill[gray!80] (2,4) rectangle (3,5);
        \fill[gray!80] (0,3) rectangle (1,4); \fill[gray!80] (1,3) rectangle (2,4); \fill[gray!80] (2,3) rectangle (3,4); \fill[gray!80] (3,3) rectangle (4,4);
        
        \foreach \y in {0,1,2} {
            \foreach \x in {0,1,2,3} {
                \fill[gray!80] (\x,\y) rectangle (\x+1,\y+1);
            }
        }
        
        \fill[gray!80] (4,2) rectangle (5,3); 
        \fill[gray!80] (5,1) rectangle (6,2); 
        \fill[gray!80] (6,0) rectangle (7,1); 
        
        \draw[thick] (0,0) rectangle (7,7);
        \draw[thick, black] (0,3) -- (7,3); 
        \draw[thick, black] (4,0) -- (4,7); 
        
        \node[rotate=90, anchor=center] at (-0.8, 5) {\small History};
        \node[rotate=90, anchor=center] at (-0.8, 1.5) {\small Queries};
        
        \node[anchor=center] at (2, 7.6) {\small History};
        \node[anchor=center] at (5.5, 7.6) {\small Queries};
        
    \end{tikzpicture}
    \caption{\textbf{Independent Query Attention Mask.} An example for a 7 token image ($t=4$ committed tokens and $K=3$ candidate queries). Gray cells ($\mathbf{A}_{i,j}=1$) indicate allowed attention; white cells ($\mathbf{A}_{i,j}=0$) are masked. The bottom-right block is strictly diagonal, preventing queries from attending to each other.}
    \label{fig:attn_mask}
\end{figure}

\subsection{Optimized KV Caching via Overwriting}
\hypertarget{secappkvcache}{}\label{subsec:app_kvcache}

Standard KV caching appends new tokens to the end of the cache at every step. In our framework, we evaluate $K$ temporary queries but only commit a subset (e.g., 1 position-content pair of tokens) to the sequence. We employ a cache overwriting strategy that reuses memory indices.

\subsubsection{Context Consistency.} 
The primary goal of this optimization is to ensure the model attends to a clean context. During training, the model sees a sequence of image tokens interleaved with their corresponding position token. If we simply appended all $K$ evaluated queries to the cache, the context for the next step would be polluted with $K-1$ rejected positions. This would shift the distribution away from training conditions. By overwriting the cache indices of the queries with the token(s) for the selected position(s), we ensure the next generation step attends only to the committed path, preserving the structure of the context.

\subsubsection{Memory Complexity.} 
Without overwriting, the cache would accumulate every candidate ever evaluated. In a generation process of length $N$ where the number of candidates $K_i$ decreases as the image fills up (i.e., $K_i \approx N-i$), the total cache size would grow to $\sum_{i=0}^{N} (N-i) \approx O(N^2)$. 
By overwriting the temporary queries with the committed tokens, the cache grows linearly with the sequence length $2N$ (considering interleaved position-content pairs), maintaining a memory complexity of $O(N)$.

\subsection{ImageNet Experimental Setup}
\hypertarget{secappinsetup}{}\label{subsec:app_in_setup}
\subsubsection{Base Model.} All experiments use a RandAR-XL~\cite{pang2024randar} model. We use the publicly available pretrained checkpoint as our any-order model.

\subsubsection{Tokenizer.} We use a pre-trained VQGAN tokenizer with a codebook size of $16,384$, an embedding dimension of $8$, and a downsampling factor of $f=16$. For a $256 \times 256$ input image, this yields the $256$ discrete tokens used by the autoregressive model.

\subsubsection{Training.} For the self-distillation phase, we fine-tune the pre-trained any-order model using the extracted specialized orders. We use the AdamW optimizer with $\beta_1 = 0.9$, $\beta_2 = 0.95$, and a weight decay of $0.05$. To ensure training stability, we apply gradient clipping with a maximum norm of $1.0$. For the fine-tuning annealing experiments, we train on the ImageNet training set for a total of 15000 steps with a global batch size of 768 ($\approx$ 9 epochs). For experiments involving order annealing, we set the annealing start and end to 1 and 3 epochs, respectively. For the non annealing experiments, we train for a total of 5000 steps with a global batch size of 768 ($\approx$ 3 epochs). The learning rate is set to $1 \times 10^{-5}$ with a cosine decay schedule without a warm-up phase.

\subsubsection{Evaluation.} We report Fréchet Inception Distance (FID), Inception Score (IS), Precision, and Recall on the ImageNet~\cite{deng2009imagenet} validation set by generating 50000 samples.

\subsubsection{Hardware}
For Imagenet experiments, we use compute nodes with 4$\times$Nvidia H100 SXM5 (80 GB memory) and 32 AMD EPYC 9454 CPUs with 128G of RAM.

\subsection{Qualitative samples}
\begin{figure*}
    \centering
\includegraphics[width=0.161\linewidth,keepaspectratio]{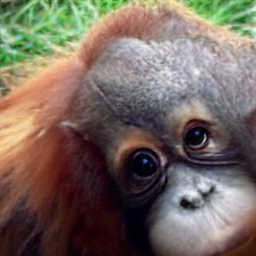}
\includegraphics[width=0.161\linewidth,keepaspectratio]{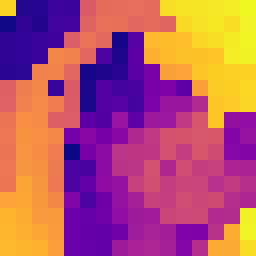}
\includegraphics[width=0.161\linewidth,keepaspectratio]{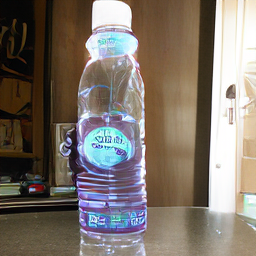}
\includegraphics[width=0.161\linewidth,keepaspectratio]{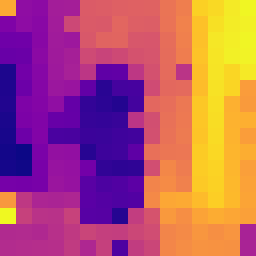}
\includegraphics[width=0.161\linewidth,keepaspectratio]{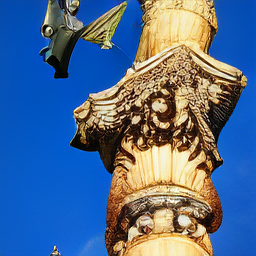}
\includegraphics[width=0.161\linewidth,keepaspectratio]{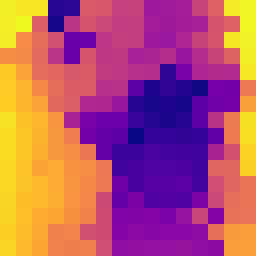}

\includegraphics[width=0.161\linewidth,keepaspectratio]{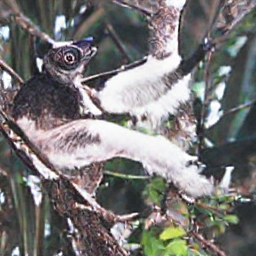}
\includegraphics[width=0.161\linewidth,keepaspectratio]{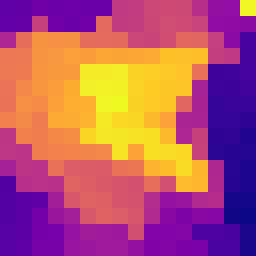}
\includegraphics[width=0.161\linewidth,keepaspectratio]{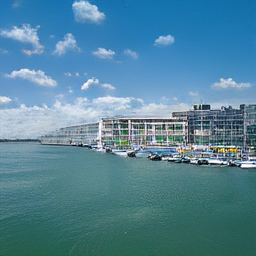}
\includegraphics[width=0.161\linewidth,keepaspectratio]{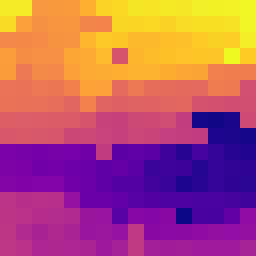}
\includegraphics[width=0.161\linewidth,keepaspectratio]{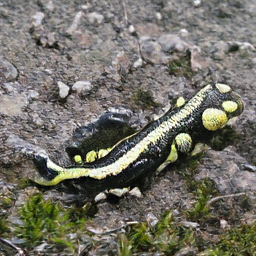}
\includegraphics[width=0.161\linewidth,keepaspectratio]{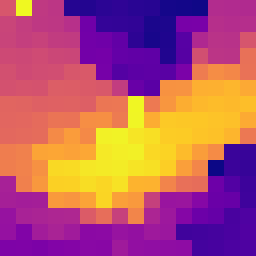}

\includegraphics[width=0.161\linewidth,keepaspectratio]{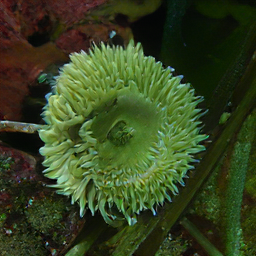}
\includegraphics[width=0.161\linewidth,keepaspectratio]{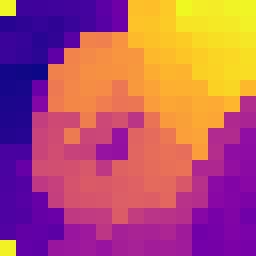}
\includegraphics[width=0.161\linewidth,keepaspectratio]{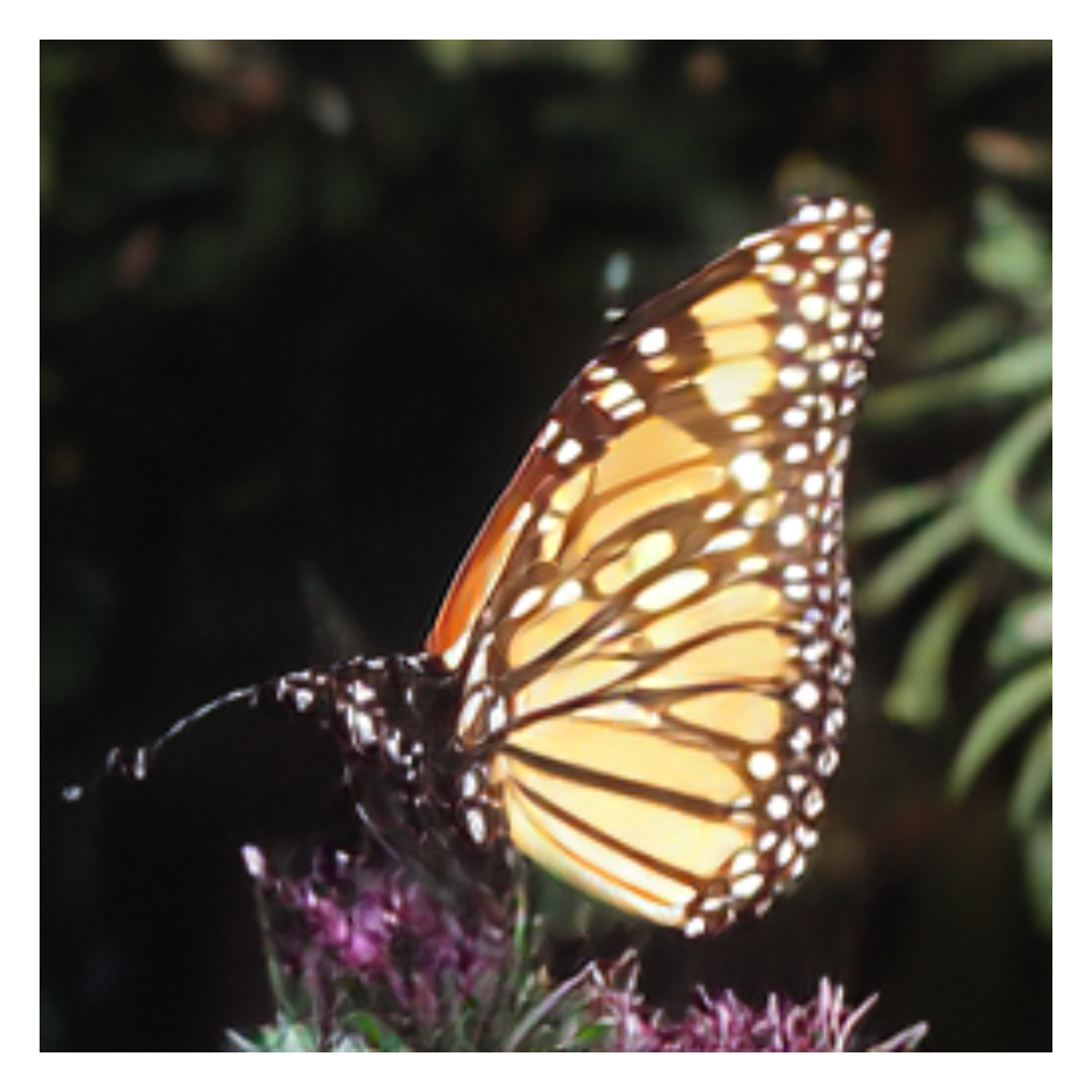}
\includegraphics[width=0.161\linewidth,keepaspectratio]{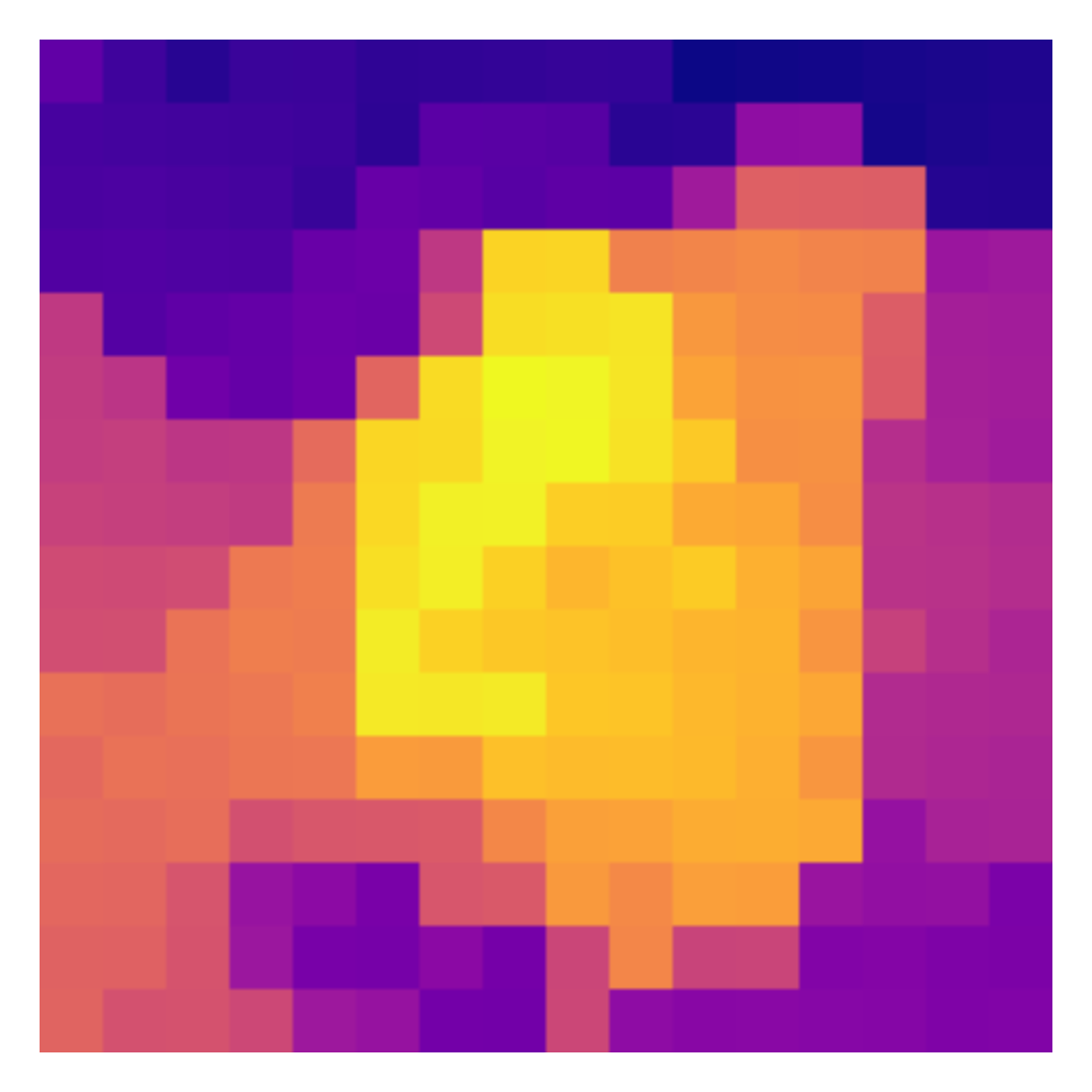}
\includegraphics[width=0.161\linewidth,keepaspectratio]{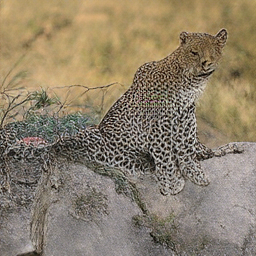}
\includegraphics[width=0.161\linewidth,keepaspectratio]{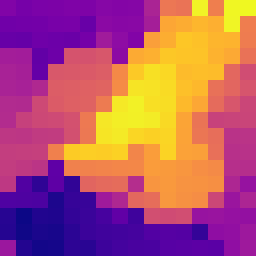}

\includegraphics[width=0.161\linewidth,keepaspectratio]{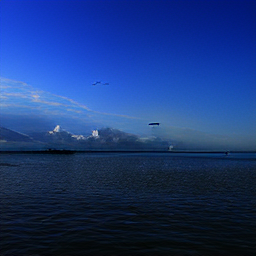}
\includegraphics[width=0.161\linewidth,keepaspectratio]{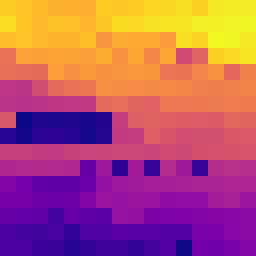}
\includegraphics[width=0.161\linewidth,keepaspectratio]{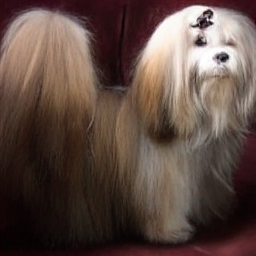}
\includegraphics[width=0.161\linewidth,keepaspectratio]{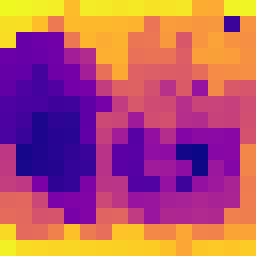}
\includegraphics[width=0.161\linewidth,keepaspectratio]{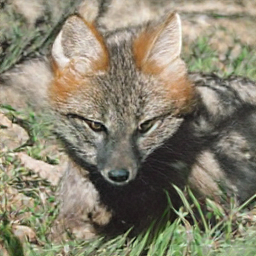}
\includegraphics[width=0.161\linewidth,keepaspectratio]{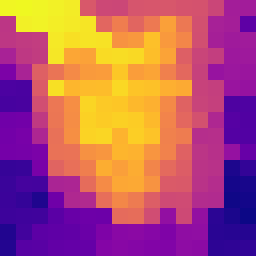}

\includegraphics[width=0.161\linewidth,keepaspectratio]{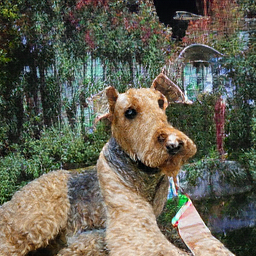}
\includegraphics[width=0.161\linewidth,keepaspectratio]{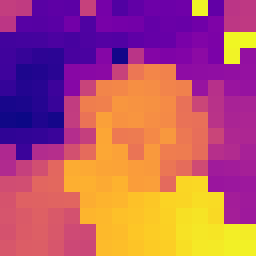}
\includegraphics[width=0.161\linewidth,keepaspectratio]{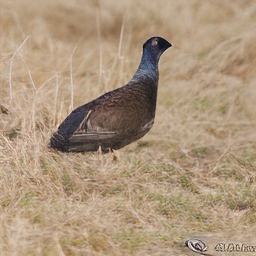}
\includegraphics[width=0.161\linewidth,keepaspectratio]{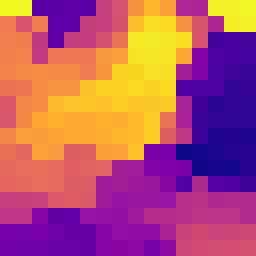}
\includegraphics[width=0.161\linewidth,keepaspectratio]{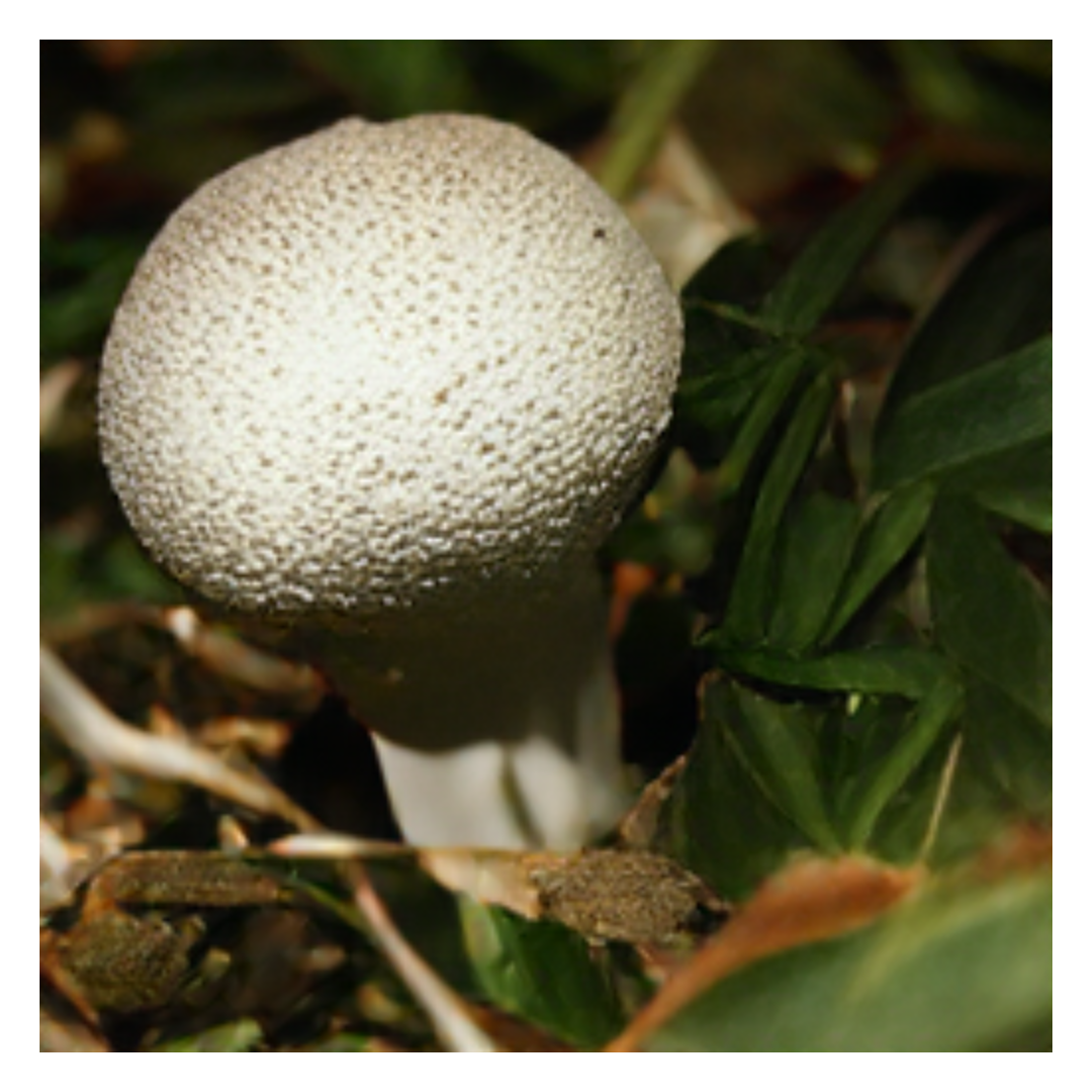}
\includegraphics[width=0.161\linewidth,keepaspectratio]{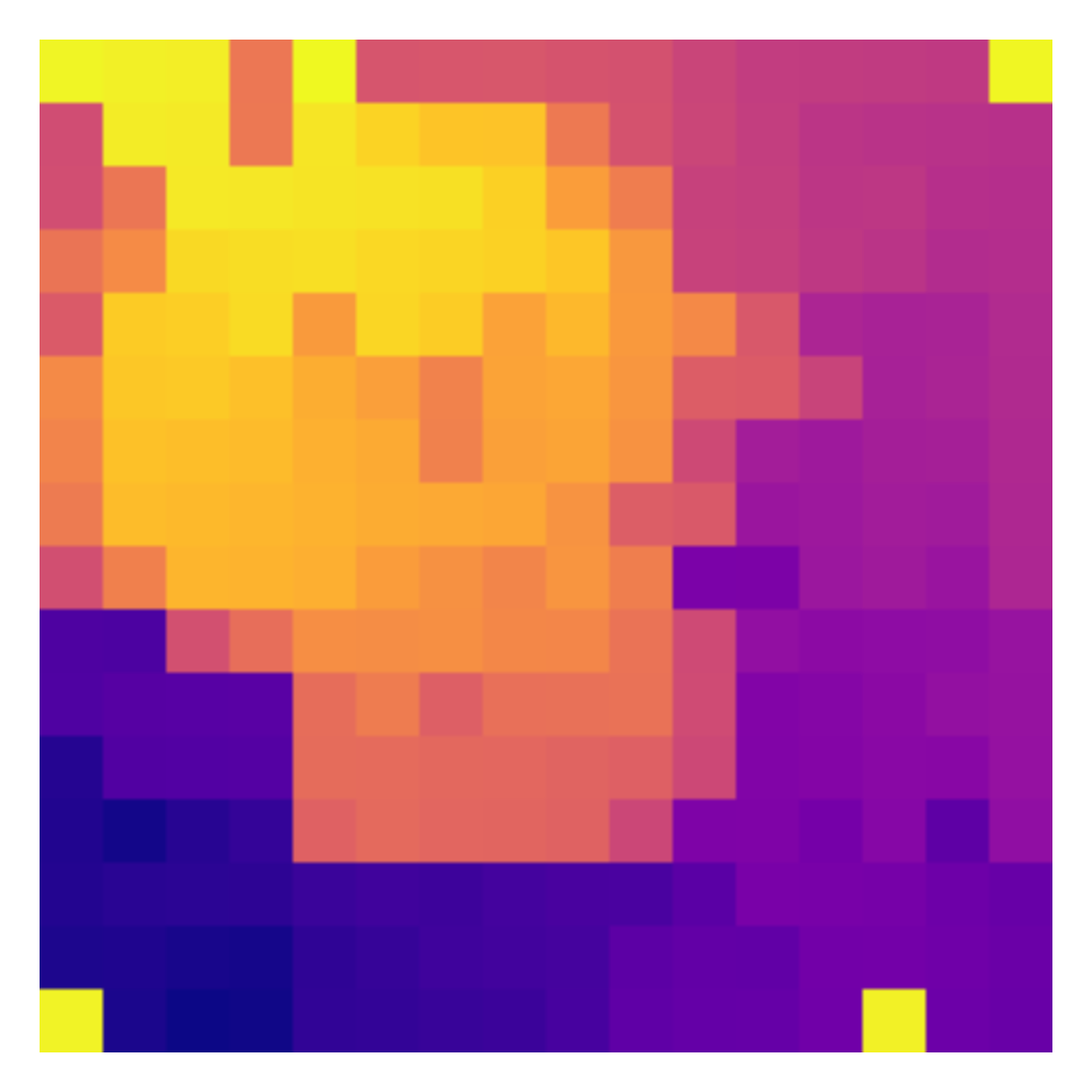}

    \caption{\textbf{Visualization of Images and Specialized Orders.} Examples of generated images with corresponding generation order heatmaps. Lighter areas represent patches generated earlier by the model. The visualization shows that the model does not follow a fixed heuristic (e.g., always background first). After self-distillation, the model prioritizes high-confidence regions---whether homogeneous backgrounds (e.g., clear skies) or salient features (e.g., the goldfish)---before tackling more ambiguous regions.}
    \label{fig:imagenet-e}
\end{figure*}

\hypertarget{secappinqual}{}\label{subsec:app_in_qual}
We present some additional examples to qualitatively demonstrate the quality of generation and the corresponding orders in Fig~\ref{fig:imagenet-e}.

\subsection{Human Evaluation Protocol, Study Design, and Results}
\hypertarget{secapphuman}{}\label{subsec:app_human}

We describe the steps we took to conduct the human evaluation study, the ethical considerations involved, and the detailed outcomes. All participants were first shown a consent statement explaining that their participation was voluntary, that they could stop at any time without penalty, and that their responses were provided freely without any external pressure or influence. The statement also clarified that the study did not involve any monetary or non-monetary compensation. Only participants who agreed to these terms were allowed to continue. Those who declined were shown a short thank-you message and exited the study immediately, and we did not collect any information from them.

Participants who consented proceeded to a series of image-pair comparison questions. In each pair, one image was generated by our OAR model and the other by the RandAR-XL baseline. To keep the evaluation fair, we randomized the left–right placement of the images for every participant. To ensure consistency, we use the same starting condition vectors, same seeds and other hyperparameters like temperature and classifier free guidance (cfg) parameter. This ensures there is no possible biases towards any specific model.

Table~\ref{tab:human_results} summarizes the per-question human preference rates across all 21 evaluation pairs sampled randomly. Each percentage represents the fraction of participants who selected the OAR-generated image as the better one for that specific pair. While preference levels vary across categories—reflecting the diversity and difficulty of the ImageNet classes—the overall trend is consistent: in most cases, participants preferred the outputs of our model over those of the baseline. A subset of questions (Q10, Q12, Q14, Q17) show preference for the baseline, reflecting the inherent difficulty of certain fine-grained ImageNet classes.

Averaged across all 21 questions, our method achieved a preference rate of \textbf{64.33\%}. This aligns with the aggregate human preference reported in the main paper and reinforces that the improvements captured by automated metrics correspond to perceptible gains in visual quality.

\begin{table}[ht!]

\centering
\scriptsize

\caption{Human preference rates (\%).}
\begin{tabular}{ccccccc}
\toprule
Q1 & Q2 & Q3 & Q4 & Q5 & Q6 & Q7 \\
64.71 & 76.47 & 88.24 & 50.98 & 64.71 & 96.08 & 94.12 \\
\midrule
Q8 & Q9 & Q10 & Q11 & Q12 & Q13 & Q14 \\
80.39 & 78.43 & 23.53 & 66.67 & 19.61 & 64.71 & 25.49 \\
\midrule
Q15 & Q16 & Q17 & Q18 & Q19 & Q20 & Q21 \\
39.22 & 62.75 & 29.41 & 82.35 & 80.39 & 80.39 & 82.35 \\
\midrule
\multicolumn{3}{c}{\textbf{Average}} & \multicolumn{4}{c}{\textbf{64.33}} \\
\bottomrule
\end{tabular}
\label{tab:human_results}
\end{table}

\begin{figure*}
    \includegraphics[width=0.245\linewidth,keepaspectratio]{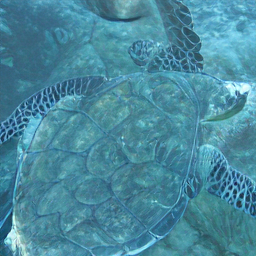}
    \includegraphics[width=0.245\linewidth,keepaspectratio]{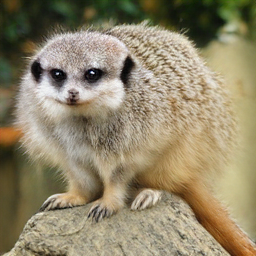}
    \includegraphics[width=0.245\linewidth,keepaspectratio]{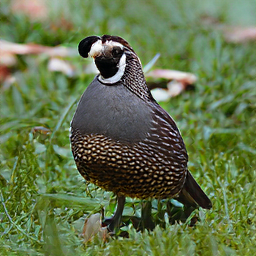}
    \includegraphics[width=0.245\linewidth,keepaspectratio]{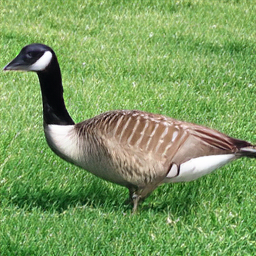}

    \includegraphics[width=0.245\linewidth,keepaspectratio]{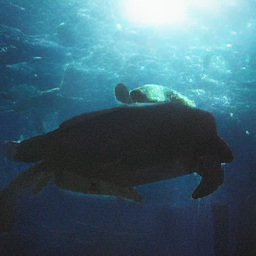}
    \includegraphics[width=0.245\linewidth,keepaspectratio]{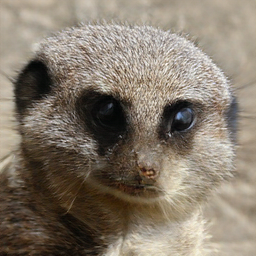}
    \includegraphics[width=0.245\linewidth,keepaspectratio]{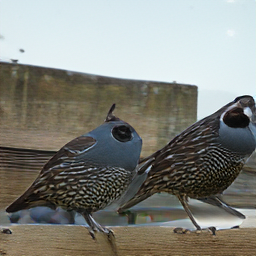}
    \includegraphics[width=0.245\linewidth,keepaspectratio]{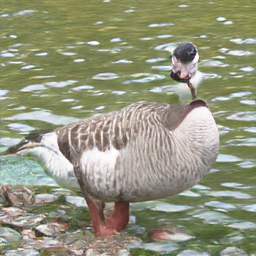}
    \caption{     
    The \textbf{top row} shows samples generated by our OAR model, while the \textbf{bottom row} displays the corresponding outputs from the RandAR-XL~\cite{pang2024randar} baseline.
    }
    \label{fig:survey_examples}
\end{figure*}
Example image pairs used in our human evaluation study are shown in Fig.~\ref{fig:survey_examples}. These pairs represent the comparison task that participants completed, where they selected whichever image they felt looked better overall, whether in terms of realism, coherence, or visual appeal.

\section{Text-to-Image}
\hypertarget{secappt2i}{}\label{sec:app_t2i}
\subsection{Generation Process}
\hypertarget{secappt2igen}{}\label{subsec:app_t2i_gen}
We provide additional details on how our approach (OAR) performs in text conditioned image generation through generation pipeline in Algorithm~\ref{alg:gene}. During inference, the transformer receives text embeddings as conditioning inputs and autoregressively predicts a sequence of discrete VQGAN~\cite{esser2021taming} token indices. To ensure we can query over all of the tokens we use a zero token padded before the first image token~\cite{pannatier2024sigma}.

To enable efficient sampling through parallelization, we use batched processing wherein each batch dimension computes one location. We use Gumbel-Top-$k$ sampling~\cite{kool2019stochastic} to sample the codebook index with the highest probability. Specifically, we use Gumbel noise~\cite{huijben2022review} with $\mu=0\;\& \;\sigma=1$.

Since we keep the pretrained VQGAN-VAE frozen, the spatial structure of the output must be restored after generation. We therefore record the model’s chosen patch locations throughout sampling and subsequently reorder the predicted tokens before decoding them into the final image. Algorithm~\ref{alg:gene} summarizes the full inference procedure.

\begin{algorithm}[ht!]
\caption{Generation Process}
\label{alg:gene}
\begin{algorithmic}[1]
\renewcommand{\algorithmicrequire}{\textbf{Input:} }
\renewcommand{\algorithmicensure}{\textbf{Output:} }
\REQUIRE Condition parameters, the generative AR transformer engine $t_\Omega$, decoder $d_\psi$
\ENSURE A generated image based on the given condition parameters
\STATE Initialize a list of possible positions $P(x,y)$ of total length $n$
\STATE Compute reference absolute $(emb^A)$ and relative $(emb^R)$ embeddings
\STATE Predict the first token using condition parameters with $t_\Omega$ 
\FOR{$i=(1,2...n)$ AR steps}
\STATE Extract the relative positions from $(P)$ for all locations
\STATE Get $emb^R$ for the relative positions
\STATE Replicate the absolute positions for $n-i+1$ times
\STATE Retrieve $emb^A$ for the absolute positions
\STATE Concatenate $emb^A$ and $emb^R$
\STATE Compute in parallel for $n-i+1$ positions 
\STATE Select the most favorable patch location (Eqn.~\ref{equ:reg})$l^*$ and record it
\STATE Use Gumbel-Top-$k$ sampling trick for the best codebook index
\STATE Remove $l^*$ from $P$
\ENDFOR
\STATE Discard the first (zero) token
\STATE Reorder the patches using recorded locations
\STATE Decode the patches tokens using $d_\psi$
\end{algorithmic}
\end{algorithm}

\subsection{V-information}
\hypertarget{secappvinfo}{}\label{subsec:app_vinfo}

In this section, we also examine the impact of order from an information-theoretic perspective~\cite{shannon1948mathematical} concerning the text conditioned image generation. When ignoring computational considerations, different factorizations of conditional probability should theoretically yield equivalent results. However, traditional information theory often overlooks essential computational factors relevant to practical applications. To address these nuances, we apply $\mathcal{V}$-information~\cite{xu2020theory}, which intuitively quantifies the additional information about a random variable $X$ that can be extracted from another variable Y using any predictor in $\mathcal{V}$. We define $\mathcal{V}$-entropy as the uncertainty we have in $Y$ after observing $X$ as follows:

\begin{align}
    H_{\mathcal{V}}(Y | X) = \inf_{f \in \mathcal{V}} E [- \log f[x](y)],
\end{align}

where $f[x](y)$ produces a probability distribution over the tokens. Information terms like $I_{\mathcal{V}}(X \rightarrow Y)$ are defined analogous to Shannon information, that is, $I_{\mathcal{V}}(X \rightarrow Y) = H_{\mathcal{V}}(Y) - H_{\mathcal{V}}(Y | X)$. In our setup, we define $\mathcal{V}$ as the Any-Order Autoregressive Model family of functions, described in Section , where $Y$ denotes the next token and $X$ represents the context tokens. These context tokens are arranged based on two distinct ordering schemes: $\mathbf{l^0}$ for the raster-scan order and $\mathbf{l^\tau}$ for the learned order inferred by our model. We define cross-entropy as $H_{\mathcal{V}}(\cdot)$, which we use to measure $\mathcal{V}$-information. Since the next tokens differ between these two orders, we calculate the sum of the $\mathcal{V}$-information from X to Y over each token of the sequence under each order and then compute the difference between them as follows:

\begin{align}
&\sum_n I_{\mathcal{V}}(\mathbf{X}_{\mathbf{l}_{<t}^\tau} \rightarrow X_{l_{t+1}^\tau}) - \sum_n I_{\mathcal{V}}(\mathbf{X}_{\mathbf{l}_{<t}^0} \rightarrow X_{l_{t+1}^0}) \\
&= \sum_n H_{\mathcal{V}}(X_{l_{t+1}^0} | \mathbf{X}_{\mathbf{l}_{<t}^0} )
- \sum_n H_{\mathcal{V}}(X_{l_{t+1}^\tau} | \mathbf{X}_{\mathbf{l}_{<t}^\tau} )
\end{align}

This difference, resulting in a value of $0.206$, quantifies the increased accessibility of $X_{l_t+1}$ when the context is arranged in the specialized order $\mathbf{l^\tau}$ as opposed to the raster-scan order $\mathbf{l^0}$ within our model family.

\subsection{Different Generation Approaches}
\hypertarget{secappgenapproaches}{}\label{subsec:app_gen_approaches}
\subsubsection{Relative position encoding}
\label{sec:rel_pos}

The simplest position encoding assigns an embedding to each absolute patch location $l_i$ $emb^A(l_i)$~\cite{vaswani2017attention}.  
However, we argue that position embedding relative to the current patch may be more effective in learning the generation of the next patch. Note that for the ImageNet experiments, we use absolute positional encodings following RandAR. For the text-to-image setting, we use two types of positional encodings in our approach for our text to image generations: the current patch position is encoded as an absolute position, while the position of the next patch is encoded relative to the current patch.
To formulate this, we compute the relative position embedding for the next patch as: $emb^R(l_{i+1})=emb^A(S+l_{i+1}-l_i)$, where $S$ is equal to the number of patches in one dimension of the image, A denotes the absolute embedding and R denotes the relative embedding. Note that parameters are not shared between the absolute and relative positional embeddings.
In relative positioning, instead of associating a position embedding $l_i$ to each patch in the image, 
we assign a positional embedding based solely on the distance of a given patch from the current position. This relative position embedding helps the model to learn to generate patches locally as it has direct access to the information on the distance between the next patch and the previous one. This setup enables the transformer to leverage both types of positional information. We use axial positional embeddings~\cite{ho2020axial} to model the positional embeddings in both cases. We use the standard image height and width for the absolute part to construct the embedding table, while for the relative part, we use double the height and width to accommodate the relative difference.
\subsubsection{Absolute and Relative encodings}

\begin{figure}[ht!]
    \centering
    \includegraphics[width=0.24\linewidth]{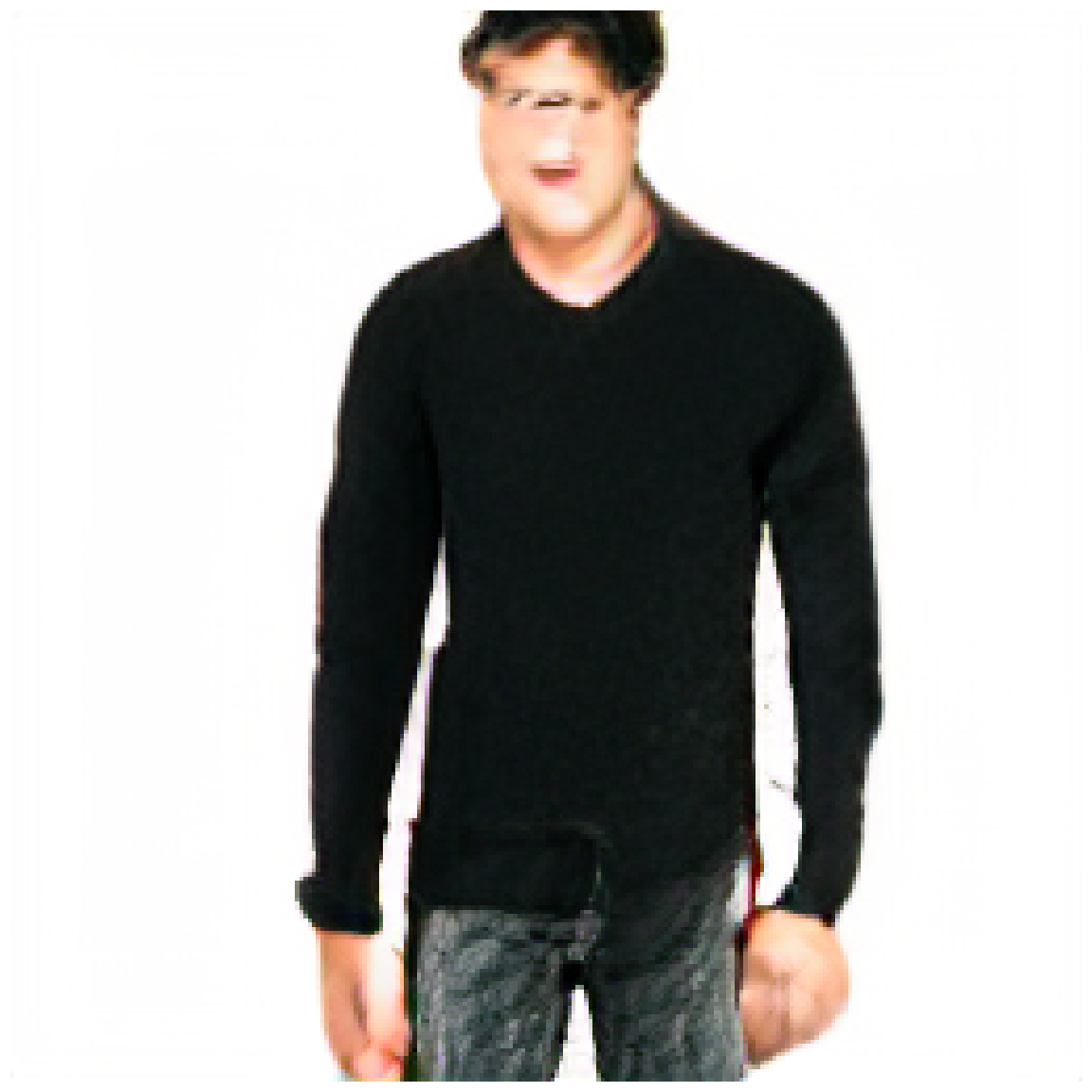}
    \includegraphics[width=0.24\linewidth]{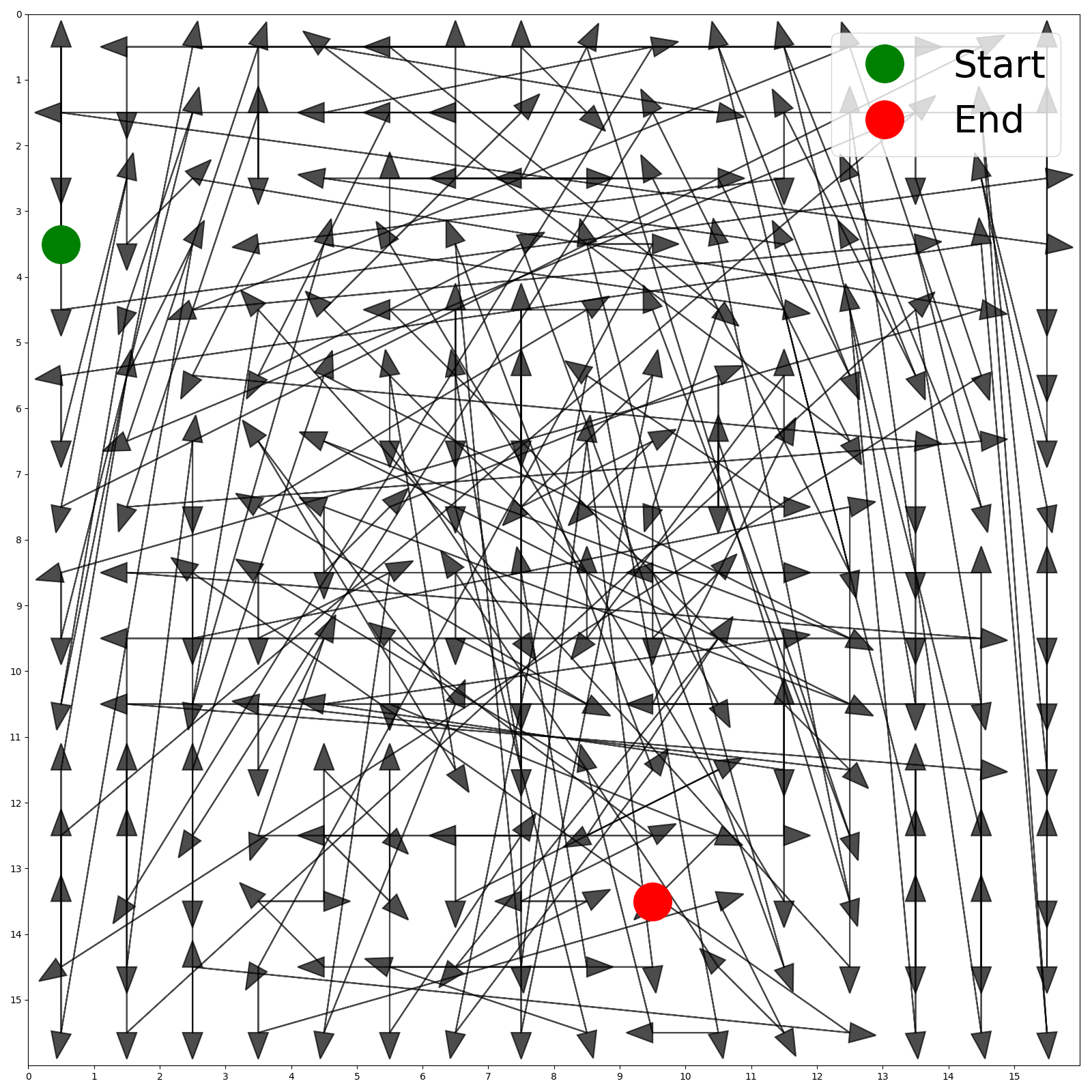}
    \includegraphics[width=0.24\linewidth]{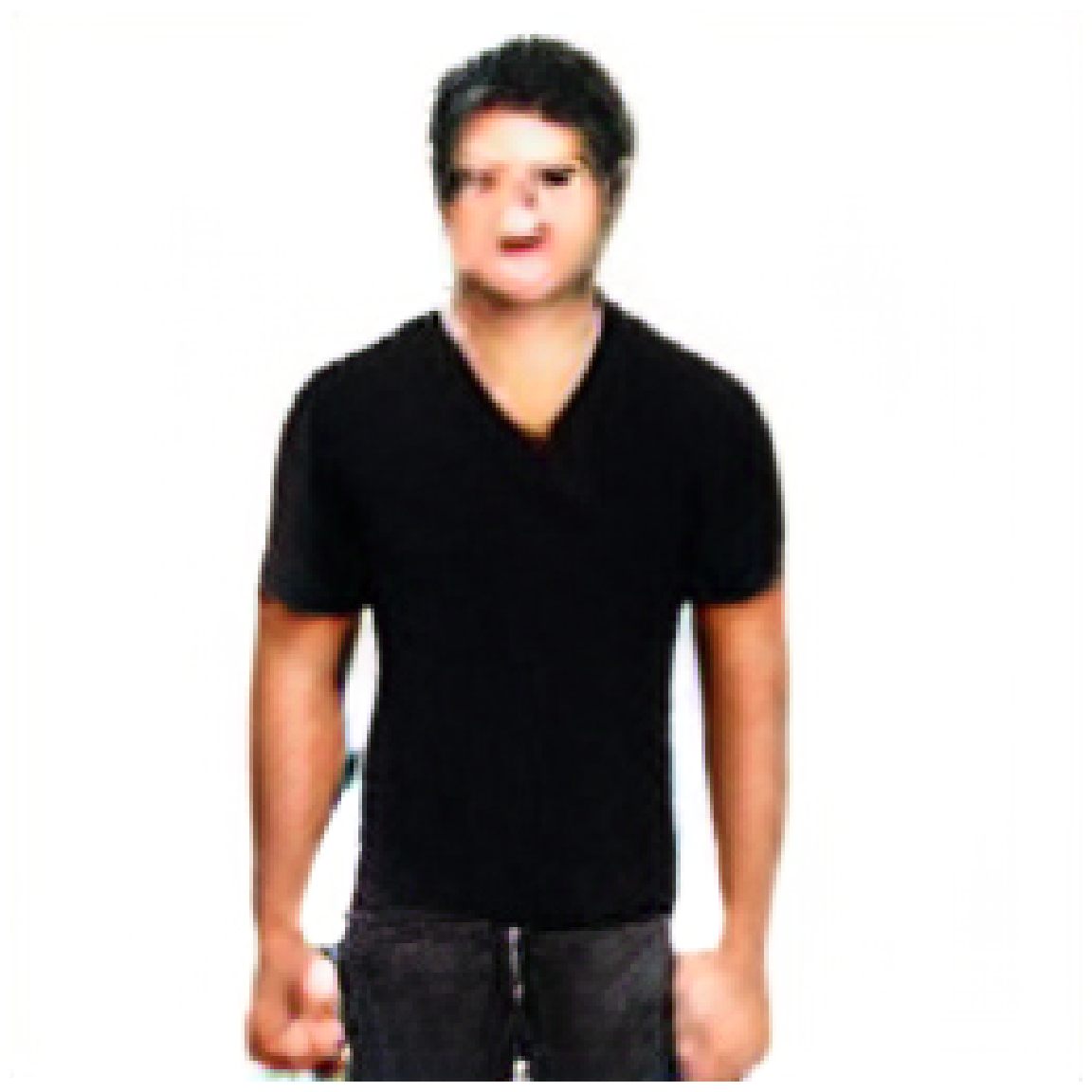}
    \includegraphics[width=0.24\linewidth]{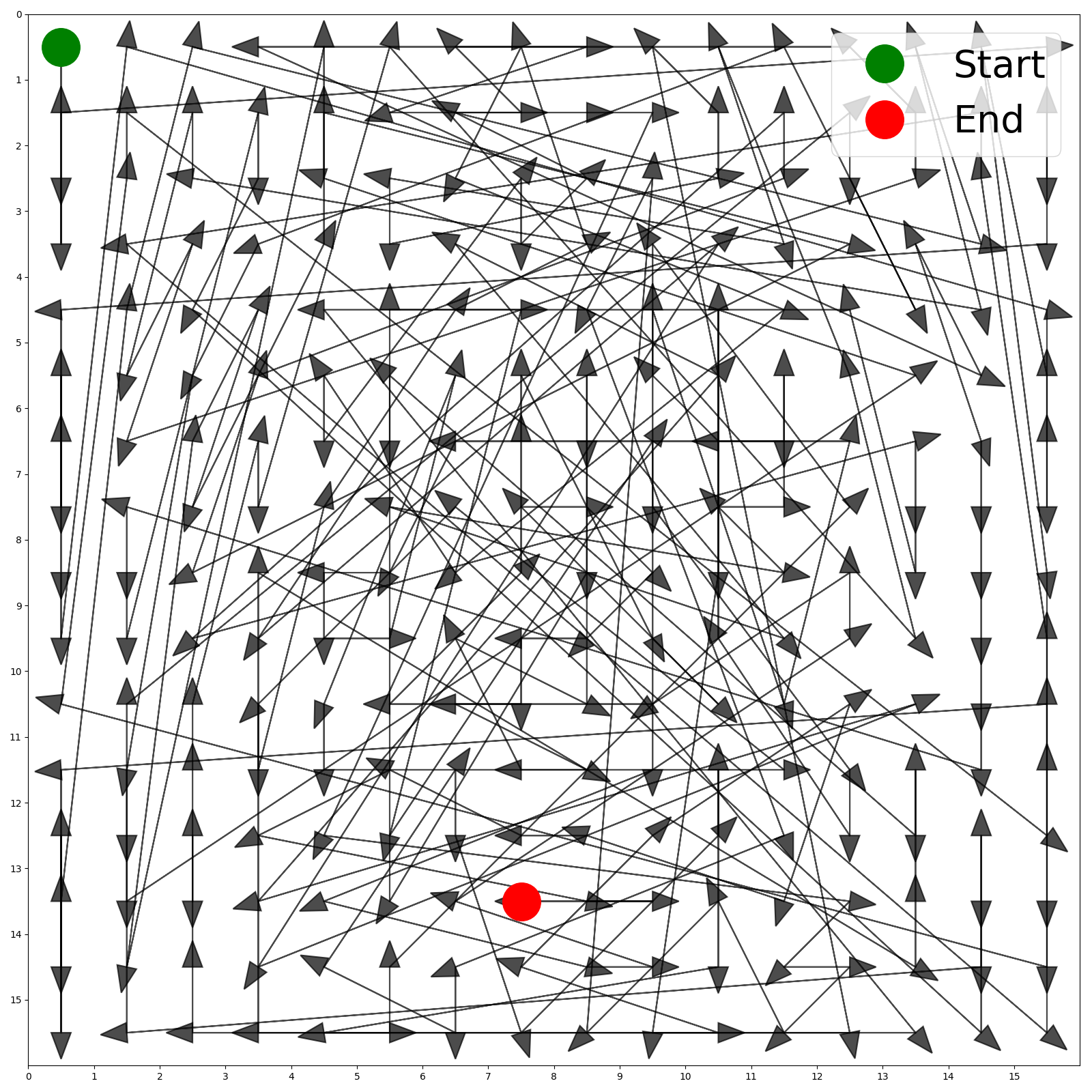}
    \caption{Generation order with absolute and relative positioning encoding. \textbf{(Left two)} With absolute encoding the generation is very scattered. \textbf{(Right two)} With relative positioning the generation is more localized. The average euclidean distance between the subsequently generated patches in case of absolute encoding is 5.78 whereas in case of relative encoding it is 4.34}
    \label{fig:abs_rel}
\end{figure}

We compare the use of absolute and relative encodings. To further investigate the practical implications, we observe how these encodings affect the output by visualizing the order. As illustrated in Figure \ref{fig:abs_rel}, consecutive tokens in the latent 1D-sequence are positioned more closely when generating the background, which aligns with the measured average distance between consecutive tokens $d$ for both models in Table \ref{tab:abs_rel}. While the metrics clearly indicate an enhancement in model performance, the visual differences remain subtle and difficult to discern.

\begin{table}[ht!]
    \centering
    \caption{Generation with absolute and relative positional encoding for the next token. }
    \begin{tabular}{l|c|ccc|c}
    \toprule
       Method & $l_{i+1}$ & FID $(\downarrow)$ & IS $(\uparrow)$  & KID $(\downarrow)$ & d\\
   \midrule
    Raster & - & 4.58 & 1.106 & 0.0031 & 1.83 \\
    OAR & Abs. &3.96 & 1.102 & 0.0024 & 5.78\\
    OAR & Rel. &3.02 & 1.108 & 0.0019 & 4.34\\
     \bottomrule
    \end{tabular}
    \label{tab:abs_rel}
\end{table}

\subsection{KV Caching}
\hypertarget{secappt2ikv}{}\label{subsec:app_t2i_kv}
In generative inference, modern LLMs and AR models~\cite{ge2024model,ramesh2021zero} present substantial latency and throughput concerns~\cite{2023keyformer}. Key-value (KV) caching significantly reduces the computational cost required by storing the key and value projections of the previously generated tokens to avoid re-computing them later. We should also note that the use of KV caching is mostly enabled by causal masking. In contrast, KV caching may not be implemented for bidirectional attention or other attention strategies~\cite{chang2022maskgit,li2024autoregressive} since the token sequence is not generally fixed.

In the same way, it is not straightforward to use KV caching in our method since the next token position in the sequence is not known in advance. To address this gap and speed up the inference, we remodel the traditional KV caching method~\cite{pope2023efficeiently}. 
As discussed thoroughly in the paper, our method relies on querying every available location at each AR step and generating a patch at the best one only. To accelerate this inference process, we compute the possible locations in parallel, thereby reducing the overall computation time.

We consider having a temporary secondary cache at each AR step, to store KV representations for all possible locations. Upon selecting the final location, we update the primary cache with the appropriate KV representation from the secondary cache. Once we select the desired location, the secondary KV cache is deleted to free up memory. 
We provide the broad-level implementation in Algorithm~\ref{algo:caching}.

\begin{algorithm}
\caption{Caching Strategy for text conditioned image processing}
\label{algo:caching}
\begin{algorithmic}[1]
\renewcommand{\algorithmicrequire}{\textbf{Input:} }
\renewcommand{\algorithmicensure}{\textbf{Output:} }
\REQUIRE Token sequence of length $i$, list of available locations \textbf{$l$}
\ENSURE Updated primary cache with selected key-value pair
\STATE Compute $k_i^l$ and $v_i^l$ for all $l \in \text{available\_locations}$ in parallel
\STATE Store all $k_i^l$ and $v_i^l$ in secondary cache
\STATE Select location $l^*$
\STATE Store $k_i^{l^*}$ and $v_i^{l^*}$ in primary cache
\STATE Delete secondary cache
\end{algorithmic}
\end{algorithm}
\subsubsection{Impact of Improved KV Caching}
It is important to deeply understand in terms of inference timings how our optimization strategies benefit in the generation process. To broadly classify our method leverages two key strategies to deal with efficiency concerns: $i)$ Parallelizing the generation process and $ii)$ KV caching. We tabulate the inference timings for a single image generation in Table~\ref{tab:time}

\begin{table}[ht!]
    \centering
    \caption{Inference Time Comparison Across Different Model Configurations}
    \begin{tabular}{c|c}
    \toprule
     Model Configuration &  Inference Time (sec)\\
     \midrule
     Raster Scan & 6.33\\
     Raster Scan (Cached) & 2.83\\
     \midrule
     OAR (Naïve) & 262.26\\
     OAR (Parallel Evaluation) & 73.76\\
     OAR (Parallel Evaluation + Optimized KV Cache) & 2.99\\
     \bottomrule
    \end{tabular}
    
    \label{tab:time}
\end{table}
We observe that parallelizing the generation process across different locations reduces latency by approximately $3.5\times$. Our optimized KV caching scheme accelerates inference by nearly $90\times$. With these optimizations, our model achieves inference efficiency comparable to traditional AR models that follow a raster-scan approach for a single image. However, it is to be noted that the reported timings are summed over all the steps in model inference and do not include overhead like loading the model, setting up caches etc.
\subsection{Effect of Background}
\begin{figure}[ht!]
    \centering
        \includegraphics[width=0.22\linewidth,keepaspectratio]{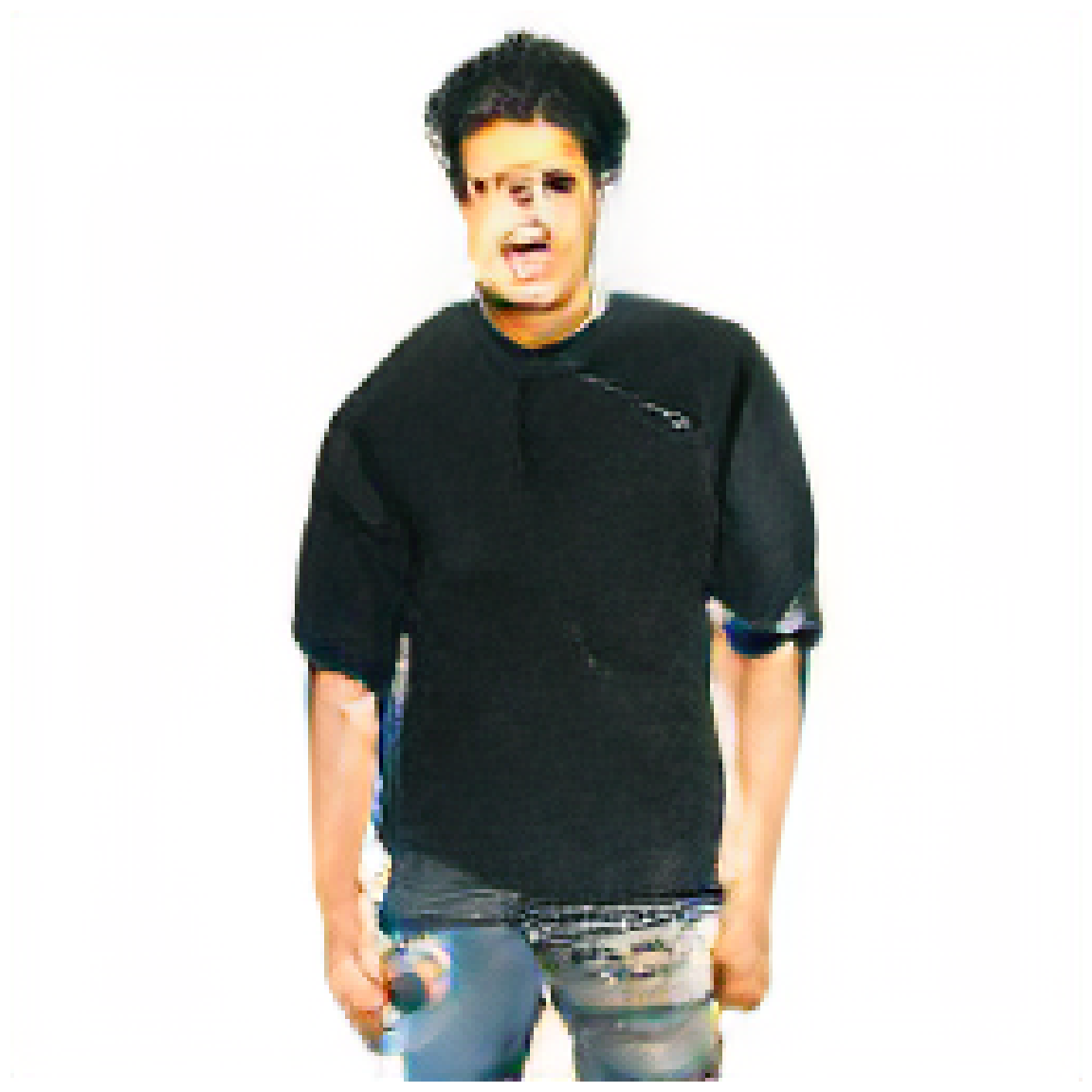}
        \includegraphics[width=0.22\linewidth,keepaspectratio]{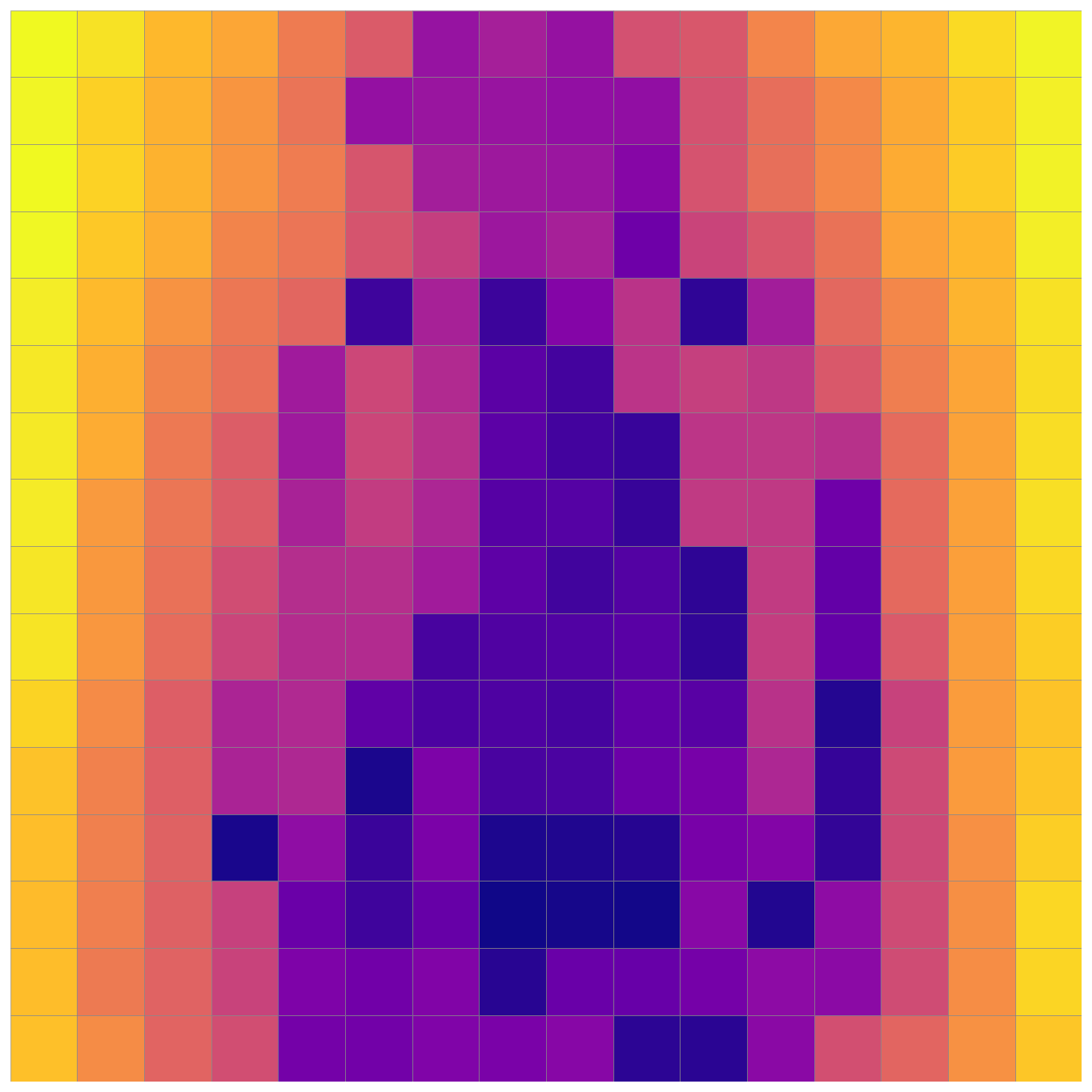}
         \includegraphics[width=0.22\linewidth,keepaspectratio]{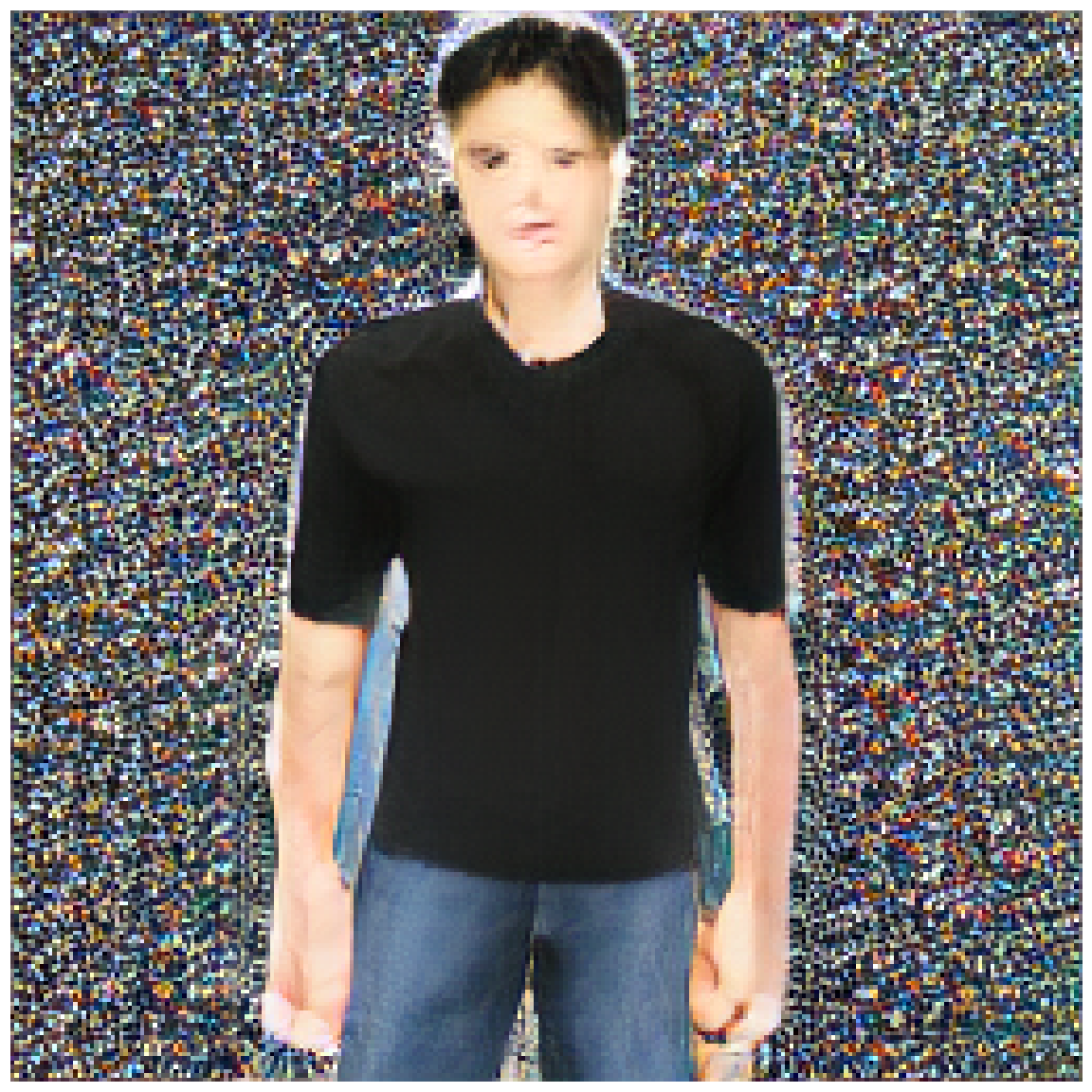}
         \includegraphics[width=0.22\linewidth,keepaspectratio]{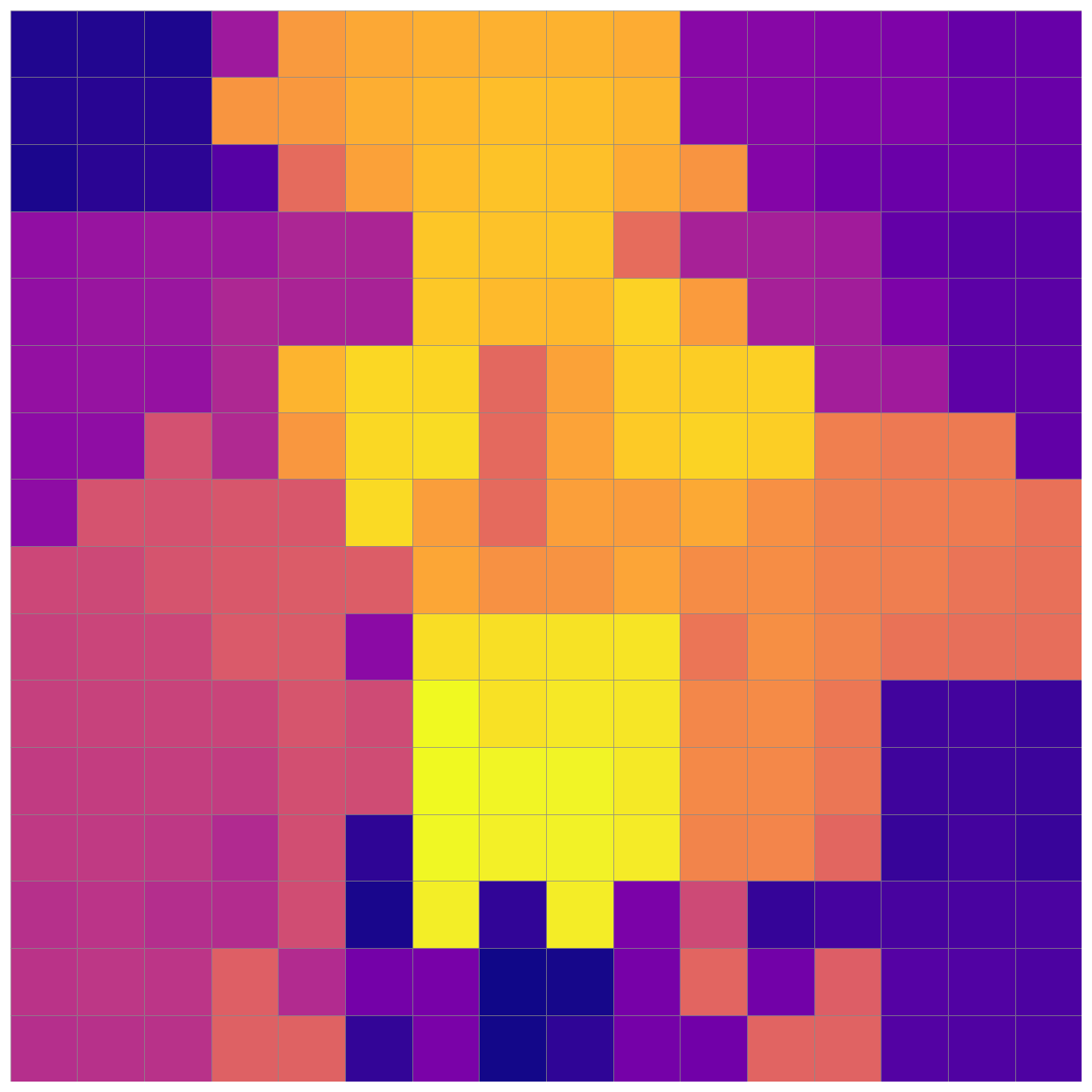}
    \caption{Generation order with different backgrounds. \textbf{(Left two)} With an easy background, such as uniform white, the model will start generating from the background. \textbf{(Right two)} With a more difficult background, such as random white noise, the model will start generating from the object. }
    \label{fig:background}
\end{figure}

\hypertarget{secappbg}{}\label{subsec:app_bg}
\begin{figure}[ht!]
    \centering
    \includegraphics[width=0.6\linewidth,keepaspectratio]{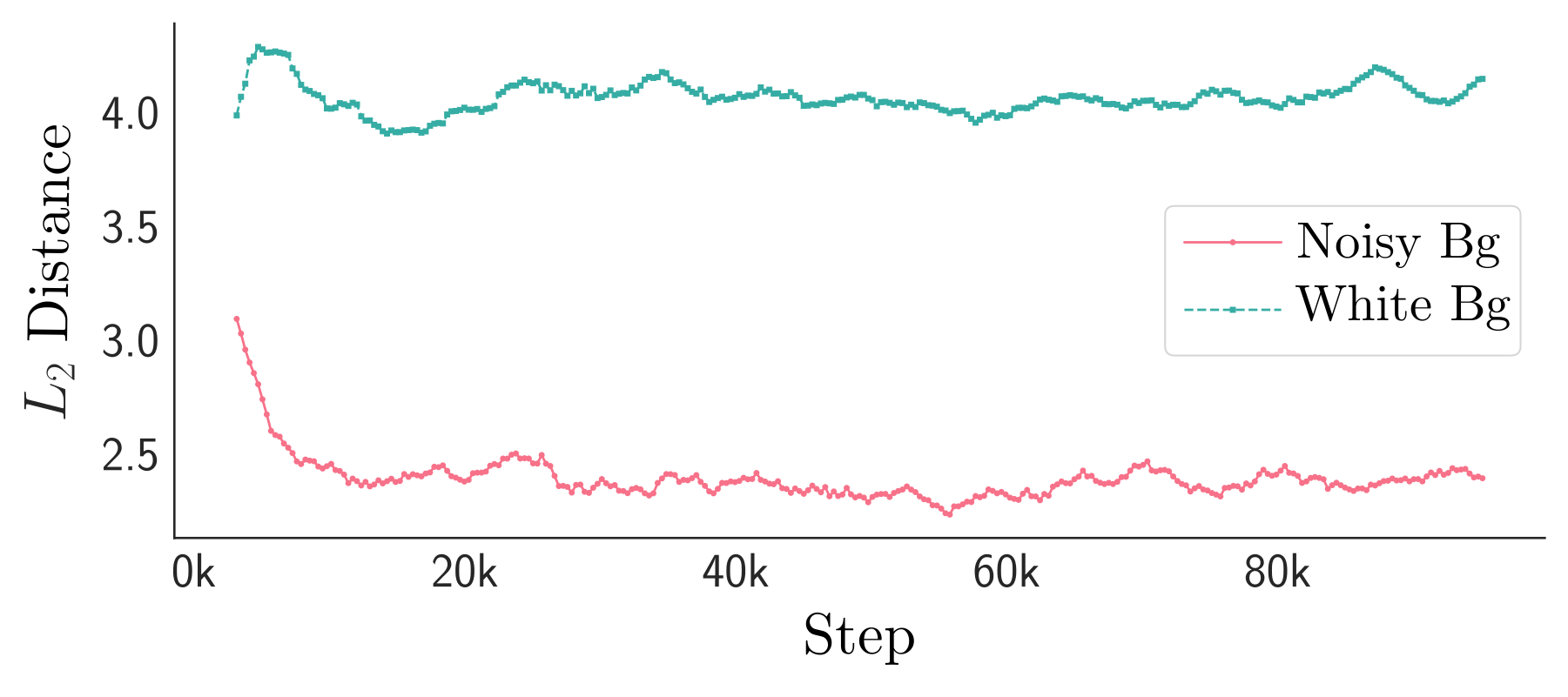}
    \caption{Average distance between generated patches for normal Fashion with white background and our modified version with a noisy background. Due to the different generation orders, the average distance is also different. }
    \label{fig:distances}
\end{figure}
All images in the dataset we used for this study have a large portion covered by white background. While initially, this could be considered an interesting feature to isolate the foreground object, in practice, it introduces biases in the generation process. 
First, the most frequent patches are the white background patches. 
Thus, those patches are the most likely to be generated and, therefore, they will be the first to be generated even though they do not contain any important information about the object of interest (see. Fig.\ref{fig:background}~(Left two)).
To verify that our model can learn orders that prioritize the foreground object, we run an additional training with a modified version of the same dataset, in which the background is filled with random noise instead of white. As the noise varies, the model cannot fit it well, so it would not be as dominant as the white background. In this setting (Fig.\ref{fig:background}~(Right two)), we observe that the model generates first the object of interest and later the background as expected. Additionally, visually, the generation seems to be improved, as the model can start the generation from the most discriminative part of the image. We tried to measure FID for this setting. Still, unfortunately, inception does not provide meaningful scores for images with random noise, so the estimated FID does not correlate well to estimate the quality of the generated images. 

Another interesting variable is the average distance $(d)$ between generated patches. 
Without inducing an external regularization (\ref{equ:reg}), we checked the evolution of the average distance during training. Results are reported in Fig.\ref{fig:distances} for the normal dataset and the variation with noisy background. In both cases, the average distance starts from around $3.5$. 
However, when the background is white, the model generates the background first. This increases the average distance to around $4.0$. In contrast, when the background is random noise, the model can focus on the foreground first, and the average distance is reduced to around $2.5$ over time. This is because when the model focuses on the object, it learns from data that locality is useful for better generation, and therefore, it learns to generate more local patches.


\subsection{Distance Regularization}
\hypertarget{secappdist}{}\label{subsec:app_dist}

To encourage smoother and more locally coherent generation orders, we add a distance-based regularization term during the selection of the next patch location. At generation step $i$, the model evaluates all possible patch locations $l$ and selects the next location $l_i^*$ according to
\begin{equation}
   l_i^* = \arg\max_l \Big( p_\theta(l, c_{i,l} \mid \mathbf{x}_{<i}, \mathbf{l}_{<i}) - \lambda\, d(l, l^{*}_{i-1}) \Big),
   \label{equ:reg}
\end{equation}
where:
\begin{itemize}
    \item $p_\theta(l, c_{i,l} \mid \mathbf{x}_{<i}, \mathbf{l}_{<i})$ is the model's predicted score (or likelihood) for choosing location $l$ at step $i$, given previously generated patches $\mathbf{x}_{<i}$ and previously selected locations $\mathbf{l}_{<i}$;
    \item $c_{i,l}$ denotes the predicted content token at location $l$ at step $i$;
    \item $l_{i-1}$ is the location chosen at the previous step;
    \item $d(l, l^{*}_{i-1})$ is the $\mathcal{L}_{\infty}$ distance between location $l$ and the previous selected location $l_{i-1}$;
    \item $\lambda$ is a regularization weight controlling the strength of the distance penalty.
\end{itemize}

This formulation introduces a trade-off between selecting the location with the highest model likelihood and staying spatially close to the previously generated patch. Larger values of $\lambda$ encourage more local, contiguous generation, whereas smaller values of $\lambda$ allow the model to prioritize high-likelihood patches even if they are spatially distant.

Table \ref{tab:reg} illustrates the impact of different values of $\lambda$, which control the trade-off between the patch likelihood and the distance from the previous token, on generation performance. As observed, the impact of this regularization is minimal. We believe this limited effect is due to the regularization being applied only at generation time, without affecting the learning. We leave the application for similar regularization during training for future work.
\begin{table}[ht!]
    \centering
    \caption{Generation with different penalty regularization.}
    \begin{tabular}{l|ccc|c}
    \toprule
    $\lambda$ & FID $(\downarrow)$ & IS $(\uparrow) $ & KID $(\downarrow)$& d\\
   \midrule
    0.0 & 3.15 & 1.106 & 0.0023& 5.34\\
    0.3 &3.11 & 1.106 & 0.0021& 4.65\\
    0.5 &3.02 & 1.108&  0.0019& 4.34\\
    0.7 &3.05 & 1.106&  0.0019& 4.09\\
     \bottomrule
    \end{tabular}
    \label{tab:reg}
\end{table}
\subsection{Decoding strategies}
\hypertarget{secappdecode}{}\label{subsec:app_decode}
In our study, we aim to maximize the use of the representation learned by our model to enhance image generation quality. To achieve this, we perform an ablation over decoding strategies. Instead of sampling the location according to Eqn.~\ref{equ:reg}, a simple alternative consists of considering the joint representation and then taking the top-k ($50\%$  of the codebook size, in our case) elements of all the patches together. We then mask the rest of the elements to ensure they are not selected. Therefore, we see the representation as a joint representation as opposed to sampling the location followed by the content. We observe the FID to slightly drop to $3.04$, the IS to be $1.109$ and the KID to be $0.0020$.

\subsection{Impact of Self-Distillation}
\hypertarget{secappdistill}{}\label{subsec:app_distill}
To evaluate the impact of the self-distillation stage, we compare our final model against a baseline trained under the same framework but without the distillation step. Both models undergo the same total training epochs to ensure a fair comparison, with the only distinction being that our model undergoes self-distillation with extracted specialized orders during the final 150 epochs.
After 450 epochs of any-order training, the baseline model achieves an FID of 2.98. In contrast, our model, which undergoes 300 epochs of any-order training followed by 150 epochs of self-distillation, achieves an improved FID of 2.56. These results show that the self-distillation stage substantially reduces FID, highlighting its crucial role in the pipeline.

\subsection{Qualitative Samples on the Multimodal CELEBA-HQ}
\begin{figure*}[ht!]
    \centering
    \includegraphics[width=0.8\linewidth,keepaspectratio]{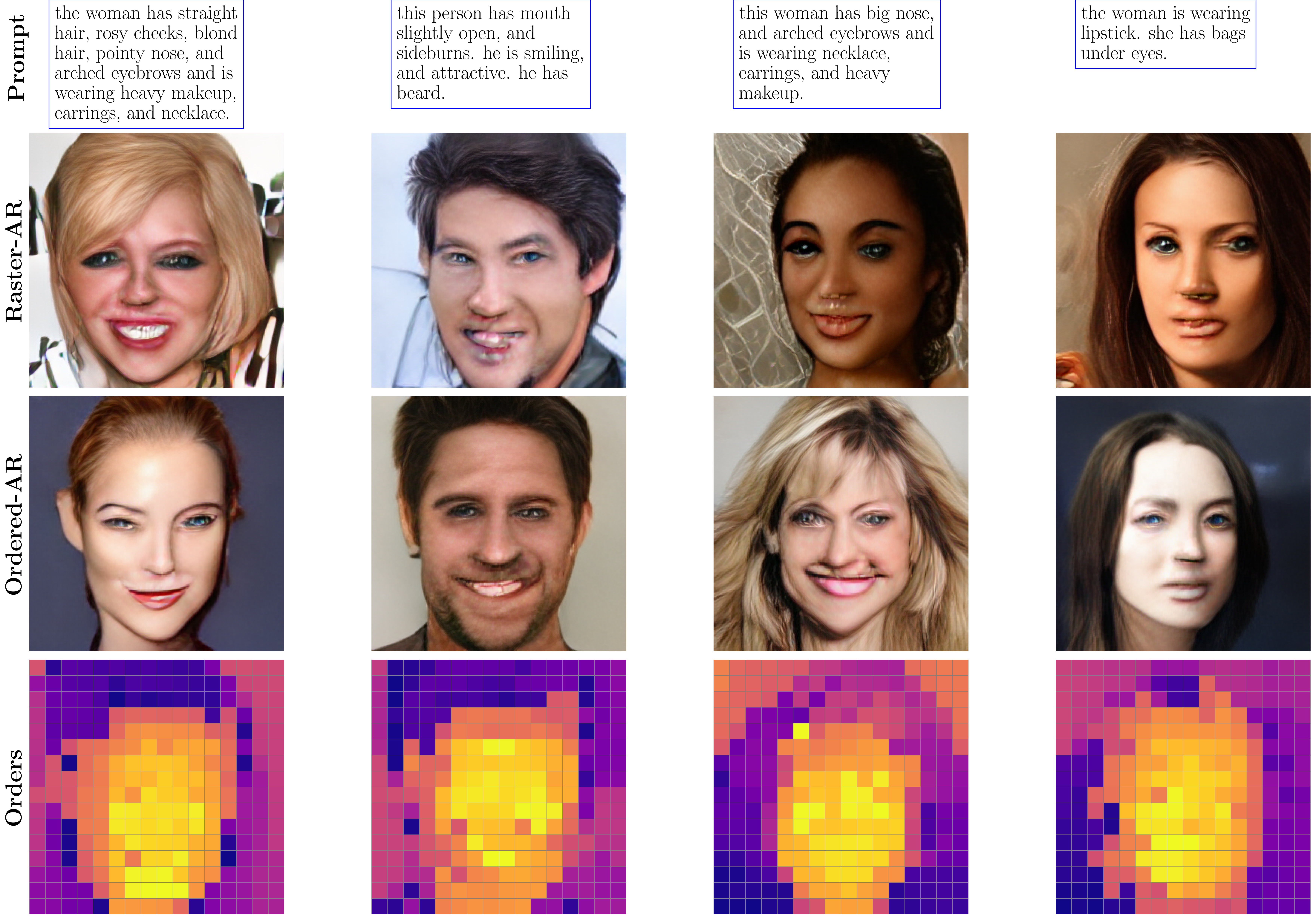}
    \caption{Examples of generation on the CelebA dataset. (Top) Generated images with raster AR mode. (Middle) Generated
images with OAR model. (Bottom) Generation order, from yellow to violet. On this dataset our model generates first the salient parts of a face, leaving hair and background at the end. Our model produces images with greater smoothness, rich context and more aligned with the text.}
    \label{fig:celeba}
\end{figure*}

\hypertarget{secappceleba}{}\label{subsec:app_celeba}
In this part of this paper we show some visual examples from the Multimodal CELEBA-HQ dataset in Fig~\ref{fig:celeba}

\end{document}